\documentclass[preprint,12pt]{elsarticle}

\usepackage[utf8]{inputenc}
\usepackage{textcomp}

\usepackage{graphicx}
\usepackage{amsmath,amsfonts,amssymb}
\usepackage{amsthm}
\usepackage[version=4]{mhchem}
\usepackage{siunitx}
\usepackage{longtable,tabularx}
\usepackage{algorithm,algorithmic}
\usepackage{caption,subcaption}
\usepackage{tikz}
\usetikzlibrary{calc,patterns,decorations.pathreplacing,calligraphy}
\tikzstyle{bag} = [align=center]

\usepackage{enumitem}

\newtheorem{proposition}{Proposition}
\newtheorem{theorem}{Theorem}

\usepackage{hyperref}

\begin{document}

\begin{frontmatter}

\title{Learning High-Dimensional Parametric Maps via Reduced Basis Adaptive Residual Networks \tnoteref{t1}\tnoteref{t2}}
\tnotetext[t1]{This work is dedicated to Professor J. Tinsley Oden in
  recognition of his lifelong seminal work in computational science
  and engineering. In particular, his foundational contributions on
  predictive modeling, uncertainty quantification, and Bayesian
  calibration \cite{BabuvskaOden04a, BabuvskaOden05, DebBabuskaOden01,
    OdenBabuvskaNobileEtAl05, OdenMoserGhattas10, Oden18,
    OdenBabuskaFaghihi17} provide motivation for the construction of the parsimonious residual neural network surrogates proposed here.}
\tnotetext[t2]{This research was partially funded by the U.S. Department of Energy under ARPA-E award DE-AR0001208 and ASCR awards DE-SC0019303 and DE-SC0021239; and the U.S. Department of Defense under MURI award FA9550-21-1-0084.}

\author[1]{Thomas~O'Leary-Roseberry\corref{cor1}}
\ead{tom@oden.utexas.edu}

\author[2]{Xiaosong Du}
\ead{xsdu@umich.edu}

\author[1]{Anirban Chaudhuri}
\ead{anirbanc@oden.utexas.edu}

\author[2]{Joaquim R. R. A. Martins}
\ead{jrram@umich.edu }

\author[1,3]{Karen Willcox}
\ead{kwillcox@oden.utexas.edu}

\author[1,4]{Omar Ghattas}
\ead{omar@oden.utexas.edu}

\affiliation[1]{organization={Oden Institute for Computational Engineering \& Sciences, The University of Texas at Austin},
            addressline={201 E 24th Street}, 
            city={Austin},
            postcode={78712}, 
            state={Texas},
            country={USA}}

\affiliation[2]{organization={Department of Aerospace Engineering, University of Michigan},
            addressline={1320 Beal Ave}, 
            city={Ann Arbor},
            postcode={48109}, 
            state={Michigan},
            country={USA}}

\affiliation[3]{organization={Department of Aerospace Engineering and Engineering Mechanics, The University of Texas at Austin},
            addressline={2617 Wichita Street, C0600}, 
            city={Austin},
            postcode={78712}, 
            state={Texas},
            country={USA}}

\affiliation[4]{organization={Walker Department of Mechanical Engineering, The University of Texas at Austin},
            addressline={204 E Dean Keeton Street}, 
            city={Austin},
            postcode={78712}, 
            state={Texas},
            country={USA}}



\begin{abstract}

We propose a scalable framework for the learning of high-dimensional parametric maps via adaptively constructed residual network (ResNet) maps between reduced bases of the inputs and outputs. When just few training data are available, it is beneficial to have a compact parametrization in order to ameliorate the ill-posedness of the neural network training problem. By linearly restricting high-dimensional maps to informed reduced bases of the inputs, one can compress high-dimensional maps in a constructive way that can be used to detect appropriate basis ranks, equipped with rigorous error estimates. A scalable neural network learning framework is thus to learn the nonlinear compressed reduced basis mapping. Unlike the reduced basis construction, however, neural network constructions are not guaranteed to reduce errors by adding representation power, making it difficult to achieve good practical performance. Inspired by recent approximation theory that connects ResNets to sequential minimizing flows, we present an adaptive ResNet construction algorithm. This algorithm allows for depth-wise enrichment of the neural network approximation, in a manner that can achieve good practical performance by first training a shallow network and then adapting. We prove universal approximation of the associated neural network class for $L^2_\nu$ functions on compact sets. Our overall framework allows for constructive means to detect appropriate breadth and depth, and related compact parametrizations of neural networks, significantly reducing the need for architectural hyperparameter tuning. Numerical experiments for parametric PDE problems and a 3D CFD wing design optimization parametric map demonstrate that the proposed methodology can achieve remarkably high accuracy for limited training data, and outperformed other neural network strategies we compared against. 


\end{abstract}



\end{frontmatter}

\section{Introduction}

We propose a scalable neural network framework for approximating high-dimensional parametric mappings by first linearly restricting the mapping to reduced bases of the inputs and outputs, and then adaptively constructing a nonlinear residual network (ResNet) approximation of the map between the reduced bases. In this framework both the \emph{compressibility} (reduced basis rank) and \emph{nonlinearity} (ResNet depth) of the map are inferred in a constructive manner, significantly reducing the need for architectural hyperparameter tuning. We denote generic parametric mappings by $m \mapsto q$, where $m \in \mathbb{R}^{d_M}$ are model parameters (input data) distributed with probability measure $m\sim \nu$ and $q \in \mathbb{R}^{d_Q}$ are quantities of interest (output data), which depend on inputs $m$. The need for parametric surrogates arises in many problems in computational science and engineering. In so called ``outer-loop'' problems, complex computational models need to be evaluated repeatedly for differing values of input parameters, making their solutions intractable if one is constrained to use only a high fidelity model. This class of problems includes uncertainty quantification, inverse problems, optimal experimental design, and optimal design and control. Another class of problems requiring accurate surrogates is real-time decision making about complex systems; this includes early warning systems, real-time inference, and real-time control.

In the emerging field of scientific machine learning (SciML) there has been a significant amount of research on surrogate strategies for learning high-dimensional parametric maps; in particular the so-called ``neural operators'' have gained significant attention due to their ability to learn high-dimensional parametric maps from data and perhaps physics informed objective functions \cite{BhattacharyaHosseiniKovachki2020,CaoOLearyRoseberryJhaEtAl22,FrescaManzoni2022,KovachkiLiLiuEtAl2021,LiKovachkiAzizzadenesheliEtAl2020a,LiKovachkiAzizzadenesheliEtAl2020b,OLearyRoseberryVillaChenEtAl2022,OLearyRoseberryChenVillaEtAl2022}. Once a sufficiently accurate surrogate has been trained, it can be substituted for the true high-dimensional map, thus making tractable the solution of outer-loop problems or real-time decision making by bypassing the full high-fidelity simulation. 

A fundamental issue exists in this setting: due to the expensive nature of evaluating high-dimensional discretized physical models, one is likely to be confronted with the difficult task of learning formally high-dimensional nonlinear maps from few training data samples. Given too many neural network weights, and too few data to learn from, one is likely to overfit to sample data and have an unreliable surrogate in practice. One strategy to overcome this lack of information is to incorporate physical principles into the training objective; this however is not always possible because in general the data may not come from a priori known conservation laws. Another strategy is to exploit compressibility of the map, if it exists. In order to avoid that the neural network weight dimension depends directly on the nominal discretization dimensions $d_M$ and $d_Q$, one can instead learn a mapping between lower dimensional reduced bases of the inputs and outputs, which can be inferred independent of the neural network training \cite{BhattacharyaHosseiniKovachki2020,OLearyRoseberryVillaChenEtAl2022}. By exploiting the structure of the input parameter uncertainty, principal information of the outputs, and derivative based sensitivity of the map (if it is available) one can bypass the high-dimensional learning problem and instead learn a lower dimensional nonlinear approximation between informed reduced bases of the inputs and outputs. This strategy is motivated by rigorous error analysis for the approximation of high-dimensional functions restricted to reduced basis mappings\cite{BhattacharyaHosseiniKovachki2020,OLearyRoseberryVillaChenEtAl2022}. This analysis helps one guide the choice of the rank for the reduced bases, thus inferring a critical neural network architectural parameter (the breadth), reducing the need to calibrate the breadth hyperparameter by ad hoc means. When high-dimensional maps can be learned in compressed representations between rank $r$ reduced bases of the inputs and outputs, this strategy reduces the neural network weight complexity from $O(d_Qd_M)$ to $O(r^2)$, which significantly reduces the complexity of the neural network training problem, and critically can help keep the number of inferred neural network weights in proportion to the amount of training data in order to help guard against overfitting. 


The reduced-basis neural network framework significantly reduces the complexity of the neural network training by use of discretization independent dimension reduction techniques, but the choice of neural network architecture for the reduced basis restricted mapping is still an open question. While the linear dimension reduction via the reduced bases represents the \emph{compressibility} of the map, the neural network architecture represents its \emph{nonlinearity}. In this work, we propose a strategy for the construction of neural network architectures that constructively detects appropriate nonlinearity for the map. We propose the use of compressed nonlinear residual network (ResNet) layers between the coefficients of the reduced input and output bases. The fundamental reason for using ResNet architectures is that they can be tweaked to add expressive power and nonlinearity without severely distorting the existing representation; this is in contrast to other neural network architectures such as the fully-connected dense networks used in \cite{BhattacharyaHosseiniKovachki2020,OLearyRoseberryVillaChenEtAl2022}. This allows one to adaptively train and construct reduced basis ResNet architectures until overfitting is detected in a unified constructive training framework. As a point of emphasis, this approach removes the need for ad hoc neural network architectural \emph{depth} hyperparameter calibration, in a similar spirit to the way the linear reduced bases remove this need for \emph{breadth}.

We motivate this algorithm and architecture by building on recent approximation theory that conceives of ResNets as discretizations of ``control flows'' \cite{LiLinShen2019}. A control flow is a dynamical system that evolves given initial conditions to match a specified target state in a pseudo-time process by minimizing a related control objective function with respect to parameters of the ODE. ResNets can be seen as a discrete analog of the control flow ODE system, where initial conditions are input data and the corresponding target state is the output data, ResNet weights are the ODE control, and the control objective is the training objective. In \cite{LiLinShen2019}, the authors are able to prove that under assumptions that are highly relevant to ResNets, control flows are universal approximators of $L^p$ functions on compact sets. The construction of their proof works by sequentially evolving local portions of input data space to their targets in output data space; arbitrary accuracy can be achieved by a finite time control flow. This theoretical construction motivates both our neural network architecture, and our construction algorithm. We use low-rank ResNet layers that are active only in subspaces of the latent space dynamics one layer at a time; the nonlinear activation function has the effect of imposing additional compression, which concentrates the effects of given layers on the latent space evolution process. The layers are constructed and trained in a sequential process as is done in the proofs of \cite{LiLinShen2019}. We prove a universal approximation property of this neural network architecture for a class of parametric maps on compact sets.

We present numerical experiments on two parametric PDE problems, and an aerodynamic inverse shape optimization problem. In the PDE problems the reduced basis strategies represent finite dimensional approximations of infinite dimensional bases related to the continuum PDE problems. In these problems, significant dimension reduction can be exposed, and the adaptive reduced basis ResNet architectures outperformed the other neural networks we considered while maintaining low dimensional neural network weights. In the aerodynamic inverse shape optimization problem, the parametric mapping represented the map from wing design and flow constraints to the constrained drag optimal wing shapes. In this case, the parameters are low dimensional, so no input dimension reduction was needed, but the use of a POD basis was able to reduce the dimensionality by $95\%$ and maintain superior performance to fully-connected encoder-decoder networks that have orders of magnitude larger weight dimensions. Numerical results demonstrate that the reduced basis adaptive ResNet strategy performs well on a variety of problems: from continuum PDE parametric maps to unstructured data coming from aerodynamic shape optimization.


\subsection{Relevant work}

A recent topic of interest in scientific machine learning is the deployment of neural networks as parametric surrogates \cite{BhattacharyaHosseiniKovachki2020,BollingerSchaeffer2020,FrescaManzoni2022,KovachkiLiLiuEtAl2021,LiKovachkiAzizzadenesheliEtAl2020b,LuJinKarniadakis2019,NelsenStuart2020,NguyenBui2021model,OLearyRoseberryVillaChenEtAl2022}. Of particular relevance to this work are projection based parametric neural surrogates which seek to parametrize high dimensional maps by use of linear and nonlinear dimension reduction strategies \cite{BhattacharyaHosseiniKovachki2020,BollingerSchaeffer2020,FrescaManzoni2022,NelsenStuart2020,OLearyRoseberryVillaChenEtAl2022}. 

Relevant to the view of the neural networks as discretizations of ODEs are the publications \cite{ChenRubanovaBettencourtEtAl2018,LiLinShen2019,RuthottoHaber2020}, which provide theoretical insight into the approximation capabilities of ResNets and relations with ODEs. The use of low rank residual network layers for non-spatial data has been proposed in various works before, but has not been widely adopted \cite{OLearyRoseberry2020,YaguchiSuzukiSNittaEtAl2019}. Adaptive training has been proposed in recent works for ResNets \cite{ChanYuYouEtAl2021,DongLiuLiEtAl2020}, and neural machine translation \cite{LiWangLiuEtAl2020} and has performed well. Distinguishing from these works, we use an approximation theoretic perspective to motivate the use of low rank ResNet layers, as well as adaptive construction and training. In our framework the low rank layers helps satisfy a requirement of the approximation theory for control flows that requires that nonlinear perturbations in ResNets work locally, and additionally the adaptive construction and training enriches the nonlinear expressive capabilities of the ResNet until an adequate approximation is obtained, all while keeping the deep neural network training problem manageable computationally.

\subsection{Contributions}

The fundamental contribution of this work is a neural network architectural framework for learning high-dimensional parametric maps via adaptively constructed reduced basis ResNets. This framework reduces the need for tuning key neural network architectural hyperparameters: breadth and depth. This framework builds on the reduced basis neural network paradigm which posits conditions for when appropriate reduced bases can be directly constructed to linearly compress high dimensional maps, which can then be nonlinearly approximated by neural networks. In this work we contribute an algorithmic framework for a class of neural networks that allow one to detect appropriate nonlinearity via the use of low rank ResNet layers, which are constructed and trained adaptively, yielding practically good performance.


The contributions of the present paper are both approximation theoretic and algorithmic. We analyze the approximation properties of this class of neural network model culminating in a universal approximation property in Theorem \ref{projected_resnet_representation_theorem}. Additionally we analyze effects of practical significance such as statistical sampling error. We then present a constructive algorithm for the reduced space adaptive ResNets, which mirrors the approximation theoretic construction of ResNets as universal approximators \cite{LiLinShen2019}. Specifically, in this work, the universal approximation property of ResNets is achieved by constructing right-hand sides of control flow ODEs that work on targeted regions of the latent space while leaving others untouched, and do so sequentially, eventually achieving a desired latent space mapping. These considerations inspire the use of low rank residual layers that work only on targeted regions of parameter space, and to construct them by the solution of sequential minimization problems. 

While neural network theory posits the rich approximation capabilities for learning complex high-dimensional maps, a gap still exists between practice and theory, in part, because the way neural networks are trained is often quite different from the way they are proven to be universal approximators of various classes of functions. The philosophy of our approach is that the construction of a neural network should attempt to mirror the theory that proves the associated classes of neural networks are universal approximators. The benefits of our approach are demonstrated for challenging parametric mapping problems that demonstrate that our targeted reduced basis ResNet methodology is capable of achieving higher accuracies for limited training data than other contemporary strategies that we compared against. A schematic for the overall approach is shown below in Figure \ref{rb_resnet_schematic}.

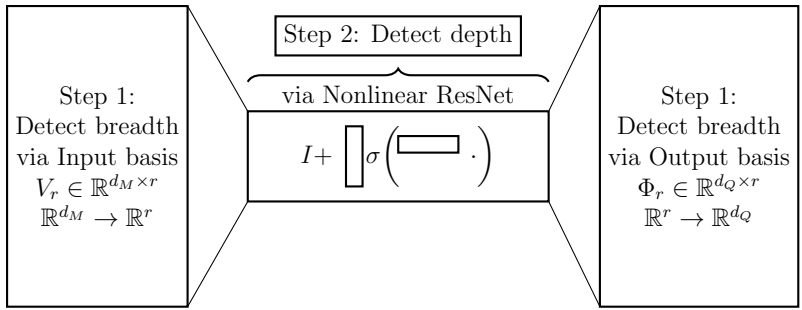
\begin{figure}[H]
\center
\begin{tikzpicture}[scale = 0.8, transform shape, every node/.style={draw,outer sep=0pt,thick}]
\node[bag] (Input) at (0,0) [minimum width=1cm,minimum height=5cm] {Step 1:\\Detect breadth\\via Input basis\\ $V_r \in \mathbb{R}^{d_M \times r}$\\ $\mathbb{R}^{d_M} \rightarrow \mathbb{R}^{r}$};
\node[bag] (NN) at (5,0) [minimum width=5cm,minimum height=1.5cm,label=via Nonlinear ResNet] {$I +$  $\quad\sigma\bigg( $  $\enskip\qquad\cdot\bigg)$};
\node[bag] (left_rank) at (4.27,0) [minimum width=0.1cm,minimum height=1cm]{};
\node[bag] (right_rank) at (5.5,0.2) [minimum width=1cm,minimum height=0.1cm]{};
\draw (Input.north east) -- (NN.north west) (Input.south east) -- (NN.south west);
\node[bag] (Output) at (10,0)[minimum width=1cm,minimum height=5cm] {Step 1:\\Detect breadth\\via Output basis\\ $\Phi_r \in \mathbb{R}^{d_Q \times r}$ \\ $\mathbb{R}^{r} \rightarrow \mathbb{R}^{d_Q}$};
\draw (NN.north east) -- (Output.north west) (NN.south east) -- (Output.south west);
\draw [decorate, thick, decoration = {brace, raise = 10pt,amplitude = 5pt}](2.5,0.8) -- (7.5,0.8) node[pos =0.5,above =25pt,black]{Step 2: Detect depth};
\end{tikzpicture}
\caption{Schematic for Reduced Basis ResNet, for a particular choice of reduced dimension.}
\label{rb_resnet_schematic}
\end{figure}

\subsection{Notation and Assumptions}

Throughout the paper we refer to $\nu$ as a probability measure, but since we work in finite dimensions we often denote the probability density function $\pi$ which we denote by the relationship $d\nu(m) = \pi(m)dm$. We consider the discrete Bochner space $L^2(\mathbb{R}^{d_M},\nu;\mathbb{R}^{d_Q})$, which we denote $L^2_\nu$ for brevity. The norm for this space is defined by its square, $f \in L^2_\nu$ if
\begin{equation}
  \|f\|^2_{L^2_\nu} = \mathbb{E}_\nu [\|f\|^2_{\ell^2(\mathbb{R}^{d_Q})}] = \int_{\mathbb{R}^{d_M}} \|f\|_{\ell^2(\mathbb{R}^{d_Q})}^2 d\nu(m) < \infty.
\end{equation}
In certain contexts, we consider other related Bochner spaces, e.g., restriction to the compact set $K \subset\subset\mathbb{R}^{d_M}$ yields the space $L^2(K,\nu;\mathbb{R}^{d_Q})$, or the restriction to the output reduced basis coefficient space yields the related Bochner space, $L^2(\mathbb{R}^{d_M},\nu;\mathbb{R}^{r})$; when we do so we will explicitly state this. At times, for brevity of notation we use $\|\cdot\|_2$ to denote the vector $\ell^2$ norm as well as the induced matrix norm on the appropriate space. When considering issues of sample complexity we denote by $\mathbb{E}_{\{m_i\sim\nu\}}$ the conditional expectation with respect to a sample set of size $N$ of independent draws $\{m_i \sim\nu\}_{i=1}^N$.

For simplicity, we consider the mapping $m\mapsto q$ to be mean-zero. This is handled in practice by setting the last layer bias to be the mean of the data so the neural network learns only the perturbations from the sample mean. We consider the map to additionally be Lipschitz continuous, i.e., there exists constant $L_q$ such that $\|q(m_1) - q(m_2)\|_{\ell^2(\mathbb{R}^{d_Q})} \leq L_q \| m_1 - m_2\|_{\ell(\mathbb{R}^{d_M})}$, and once differentiable with respect to $m$ in the context of active subspace.

\section{Reduced-Basis Parametric Surrogates} \label{reduced_basis_surrogate_section}

This section discusses approaches for approximating high-dimensional parametric functions by restricting them to (linear) reduced bases of their inputs and outputs. This section establishes the objective of the next section: to learn the compressed nonlinear map by adaptively constructed ResNets. This section reviews some work from \cite{BhattacharyaHosseiniKovachki2020,OLearyRoseberryVillaChenEtAl2022} and the related references \cite{ManzoniNegriQuarteroni2016,QuarteroniManzoniNegri2015,SchwabTodor2006,ZahmConstantinePrieurEtAl2020}. Those familiar with reduced basis neural network methodologies can skip this section. 

\subsection{Compressing high-dimensional maps with reduced bases}

Our target is learning high-dimensional parametric mappings that admit compressible representations. Such maps arise, e.g., in the discretizations of partial differential equations that have a direct dependence on a random parameter field (see e.g., \cite{GhattasWillcox2021} and the references therein). The mappings are compressible if just a limited number of basis vectors for the inputs and outputs of the map can provide a desired accuracy in approximating the map. Such compressed representations are referred to as input-output ridge functions. A ridge function is a composition of two functions $g \circ V_{r_M}^T$, where $V_{r_M}:\mathbb{R}^{d_M} \rightarrow \mathbb{R}^{r_M}$ is a linear mapping (a matrix in $\mathbb{R}^{r_M \times d_M}$), and $g:\mathbb{R}^{r_M} \rightarrow \mathbb{R}^{d_Q}$ is a measurable function. The ridge function mitigates the dependence on the input dimension $d_M$ by restricting the mapping to an $r_M$ dimensional subspace of $\mathbb{R}^{d_M}$. The dependence on the output dimension $d_Q$ can additionally be mitigated by employing a reduced basis for the output. Since we assume the map to be mean-zero, the representation is thus
\begin{equation} \label{input_output_ridge_function}
    q_{(r_M,r_Q)}(m) = \Phi_{r_Q}g_r(V_{r_M}^Tm),
\end{equation}
which mitigates high dimensional dependence ($d_M$,$d_Q$) linearly via reduced bases of dimensions ($r_M$,$r_Q$). For mappings that are not mean-zero, an affine shift is required. Given a desired tolerance $\epsilon >0$, a map $m\mapsto q$ is linearly compressible in $L^2(\mathbb{R}^{d_M},\nu; \mathbb{R}^{d_Q})$ if there exists a ridge function \eqref{input_output_ridge_function} such that
\begin{equation} \label{generic_compression_bound}
    \|q_{(r_M,r_Q)} - q\|_{L^2_\nu} < \epsilon,
\end{equation}
given $r_M \ll d_M$, or $r_Q \ll d_Q$. In \cite{BhattacharyaHosseiniKovachki2020,OLearyRoseberryVillaChenEtAl2022}, reduced basis surrogate strategies based on proper orthogonal decomposition, Karhunen Lo\'{e}ve Expansion (KLE) \cite{SchwabTodor2006}, and active subspace (AS) \cite{ZahmConstantinePrieurEtAl2020} are shown to have this compressibility, given sufficiently rapid decay on truncation errors related to these reduced bases. We review the main ideas here. 
Classically, the optimal output basis for representing the $\mathbb{R}^{d_Q}$ output of a function in $L^2(\mathbb{R}^{d_M},\nu;\mathbb{R}^{d_Q})$ is given by the proper orthogonal decomposition (POD) \cite{ManzoniNegriQuarteroni2016,QuarteroniManzoniNegri2015}, which is the eigenvector basis for the operator $\mathbb{E}_\nu[qq^T] = \Phi^{(POD}\Lambda^{(POD)} \Phi^{(POD)T}$, sorted in descending order by eigenvalues $\lambda_i, i = 1, \dots, d_Q$. Due to the Hilbert-Schmidt theorem, the rank $r_Q$ POD basis minimizes the following error 
\begin{align}
    \Phi_{r_Q}^{(POD)} = \text{argmin}_{P_{r_Q} \in \mathbb{R}^{d_Q \times r_Q}} \|q - P_{r_Q}P_{r_Q}^Tq\|^2_{L^2_\nu}\\
    \|q - \Phi_{r_Q}^{(POD)}(\Phi_{r_Q}^{(POD)})^Tq\|^2_{L^2_\nu} = \sum_{i=r_Q +1}^{d_Q}\lambda^{(POD)}_i,
\end{align}
which makes it well suited to the task of constructing an output reduced basis for $L^2_\nu$ maps. The $\ell^2(\mathbb{R}^{d_M})$ optimal reduced basis for representing the uncertain parameter is the  Karhunen Lo\'{e}ve Expansion (KLE) basis \cite{SchwabTodor2006}, which computes an eigenvector basis representation of the covariance of $\nu$,
\begin{equation}
   \mathcal{C} = \mathbb{E}_\nu\left[ (m - \mathbb{E}_\nu[m]) (m - \mathbb{E}_\nu[m])^T\right] = V^{(KLE)}\Lambda^{(KLE)}(V^{(KLE)})^T.
\end{equation} 
When the mapping $m\mapsto q$ is Lipschitz, the effects of the input parameter truncation on the output approximation can be conservatively bounded using the Lipschitz constant,
\begin{equation} \label{kle_ridge_function_bound}
    \|q - q \circ V^{(KLE)}_{r_M}V_{r_M}^{(KLE)T}\|^2_{L^2_\nu} \leq L_q^2\sum_{i=r_M+1}^{d_M} \lambda^{(KLE)}_i,
\end{equation}
which is employed in \cite{ZahmConstantinePrieurEtAl2020} Proposition 3.1 and \cite{BhattacharyaHosseiniKovachki2020} Theorem 3.5. POD and KLE are analogous to principal component analysis (PCA) on the output quantity of interest and input parameter respectively. 
When the mapping $m \mapsto q$ has one parametric derivative, active subspaces (AS) \cite{ZahmConstantinePrieurEtAl2020} can be used to construct a goal-oriented input reduced basis for the map using both derivative sensitivity and parametric uncertainty information. This is done by
construction of a eigenvector basis for the generalized eigenvalue problem
\begin{equation} \label{active_subspace}
    \mathbb{E}_\nu[\nabla q^T \nabla q]v_i = \lambda_i^{(AS)} \mathcal{C}^{-1}v_i,
\end{equation}
where $(\lambda_i^{(AS)},v_i)_{i\geq 1}$ are the generalized
eigenpairs sorted such that $\lambda_i \geq \lambda_j$ for $ i <
j$. Proposition 3.1 in \cite{ZahmConstantinePrieurEtAl2020} establishes that for Gaussian measure $\nu$, there exists a (conditional expectation) ridge function mapping $q_{r_M}^{(AS)}$ (restricting the input parameter to the $r_M$ AS basis) such that
\begin{equation}
    \|q - q_{r_M}^{(AS)}\|^2_{L^2_\nu} \leq \sum_{i=r_M+1}^{d_M} \lambda^{(AS)}_i.
\end{equation}
This bound removes the dependence on the conservative Lipschitz constant, suggesting that by incorporating the global sensitivities of the map (captured by the AS basis) one may be able to do better than truncation of the input parameter via covariance eigenvectors. This is observed numerically in \cite{OLearyRoseberryVillaChenEtAl2022} and in Section \ref{section_numerical_experiments}. We note that KLE is simpler to compute, since AS requires additional parametric derivative computation. Scalable computation of AS requires that codes be equipped with adjoint or automatic differentiation procedures; finite differencing, while always an option yields worse computational complexity. Further discussion is outside the scope of this work. We refer the reader to Appendix A of \cite{OLearyRoseberryVillaChenEtAl2022} for a discussion of scalable methods to compute the AS basis.

Approximation bounds can be derived to demonstrate the error of the approximation by reduced-basis conditional expectation ridge functions in the KLE, AS and POD bases; this is analogous to Proposition 2.2 in \cite{OLearyRoseberryVillaChenEtAl2022}, and part of the bounds in Theorem 3.5 in  \cite{BhattacharyaHosseiniKovachki2020}. Without taking into account the effects of statistical sampling error, these bounds can take the form:
\begin{subequations}
\begin{align}
  &\|q - \widehat{\Phi}^{(POD,AS)}_{r_Q}\widehat{\Phi}^{(POD,AS)T}_{r_Q}q^{(AS)}_{r_M}\|^2_{L^2_\nu} \leq 2\sum_{i=r_M+1}^{d_M}\lambda_i^{(AS)} + 2\sum_{i=r_Q+1}^{d_Q}\widehat{\lambda}_i^{(POD)} \label{ridge_bound_as}\\
  &\|q - \widehat{\Phi}^{(POD,KLE)}_{r_Q}\widehat{\Phi}^{(POD,KLE)T}_{r_Q}q^{(KLE)}_{r_M}\|^2_{L^2_\nu} \leq \nonumber \\
   & \qquad \qquad \qquad \qquad \qquad \quad \qquad 2L_q^2\sum_{i=r_M+1}^{d_M}\lambda_i^{(KLE)} + 2\sum_{i=r_Q+1}^{d_Q}\widehat{\lambda}_i^{(POD,KLE)} \label{ridge_bound_kle}.
\end{align}
\end{subequations}
In these bounds, the POD basis is not for the true mapping $q$, but for the input-truncated ridge functions $q^{(AS)}_{r_M},q^{(KLE)}_{r_M}$ respectively; this is denoted by the use of a hat ( $\widehat{\cdot}$ ). Let $q_{r_M}$ generically stand for any input-basis truncated ridge function. We can see the spectral convergence of POD for $q_{r_M}$ to POD for $q$ by the following construction. Let $e(m) = q(m) - q_{r_M}(m)$, by equations (\ref{ridge_bound_as},\ref{ridge_bound_kle}). For any $\epsilon >0$ we can find $r_M,r_Q$ large enough that the entire trailing sum is bounded by $\epsilon^2$. By making use of Jensen's, Cauchy-Schwarz, and triangle inequalities, we can bound as follows (here let $\|\cdot \|_2,\|\cdot\|_F$ denote $\ell^2$ vector norm and Frobenius matrix norm respectively):
\begin{align}
\| \mathbb{E}_\nu [qq^T] - \mathbb{E}_\nu[q_{r_M}q_{r_M}^T]\|_F &\leq \mathbb{E}_\nu \left[\|qq^T - q_{r_M}q_{r_M}^T\|_F \right] \nonumber \\
&= \mathbb{E}_\nu[\|qe^T + eq^T - ee^T\|_F] \nonumber \\
&\leq \mathbb{E}_\nu[\|qe^T\|_F + \|eq^T\|_F + \|ee^T\|_F] \nonumber \\
&= \mathbb{E}_\nu[2\|q\|_2\|e\|_2 + \|e\|_2^2] \nonumber \\
& \leq 2\sqrt{\mathbb{E}_\nu[\|q\|_2^2]}\sqrt{\mathbb{E}_\nu[\|e\|_2^2]} + \mathbb{E}_\nu[ \|e\|_2^2]\nonumber \\
& \leq 2 \|q\|_{L^2_\nu} \epsilon + \epsilon^2.
\end{align} 

Since the trailing sums in \eqref{ridge_bound_as} and \eqref{ridge_bound_kle} can be minimized by choosing $r_M = r_Q$, for simplicity we do so, and for the rest of this work we will refer to the reduced basis dimensions as $r$. For simpler notation, we use $\Phi_r$ to denote the reduced basis going forward instead of $\widehat{\Phi}_r$. Additionally, since the coefficients of a given input-output ridge function approximation are by definition truncated in the orthogonal complements of the input and output reduced bases, from this point forward we define the input-output ridge function in reduced representation. That is, $q_r:\mathbb{R}^r \rightarrow \mathbb{R}^r$, which is extended to the full spaces by composition: $\Phi_r \circ q_r \circ V_r^T:\mathbb{R}^{d_M}\rightarrow \mathbb{R}^{d_Q}$. We use $\mathbb{R}^r \ni m_r = V_r^Tm$ to represent the restriction of the input parameter to the input reduced basis.

\subsection{Effects of statistical sampling error on reduced basis approximations} \label{effects_of_subsampling_subsection}

The reduced bases discussed in the previous sections are all formally defined as expectations of symmetric positive definite operators, and in practice have to be estimated via Monte Carlo integration (except for cases where the covariance $\mathcal{C}$ of $\nu$, is known analytically by construction). We can generically state the reduced basis construction via the (possibly generalized) eigenvalue problem:
\begin{equation} \label{general_stochastic_basis_problem}
    \mathbb{E}_\nu [AA^T]\psi_i = \lambda_i B \psi_i.
\end{equation}
In the case of POD $A = q(m),B = I$, for KLE $A = m - m_0,B = I$ and in the case of AS, $A = \nabla q(m), B = \mathcal{C}^{-1}$, noting that in the case of AS the covariance $\mathcal{C}$ is usually known a priori and thus is not a source of statistical sampling error. As a first step in any of these computations there is first a Monte Carlo approximation of $\mathbb{E}_\nu[AA^T]$, i.e., given independent sample data $\{A_i = A(m_i)|m_i \sim \nu\}_{i=1}^N$ we approximate
\begin{equation}
    \mathbb{E}_\nu[AA^T] \approx \frac{1}{N}\sum_{i=1}^NA_iA_i^T.
\end{equation}
If we know a priori that we can bound sample variance as $\|\mathbb{E}_\nu[A_iA_i^T - \mathbb{E}_{\nu}[AA^T]]\|^2_2 \leq \sigma_A^2$, then we can obtain a typical $O(N^{-1/2})$ Monte Carlo bound for the matrix $\ell^2$ norm; see \ref{AAT_mc_bound}. How the $O(N^{-1/2})$ Monte Carlo errors for the operator $\mathbb{E}_\nu[AA^T]$ propagate through the solution of \eqref{general_stochastic_basis_problem} to errors in the eigenvectors is in general not known (to the best of our knowledge). Indeed, these statistical sampling errors will introduce additional errors that propagate to the trailing sums used to select rank $r$ in \eqref{ridge_bound_as}, \eqref{ridge_bound_kle}; in the presence of statistical sampling error, the choice of the rank $r$ should also depend on the amount of data used in the estimation for the basis. For the specific case of POD (and KLE when the covariance is approximated by Monte Carlo), a bound is established in (Theorem 3.4 \cite{BhattacharyaHosseiniKovachki2020}). In this case, the bound is obtained by coupling the sample approximation to the problem of finding an optimal rank $r$ reduced bases. For $m$ the rank $r$ basis optimality is measured in the $L^2(\mathbb{R}^{d_M},\nu;\mathbb{R}^{d_M})$ norm, while for $q$ the optimality is measure in $L^2_\nu$. The same approach cannot be applied to AS since it is not optimal for representing $m$ in the parameter representation, i.e. the $L^2(\mathbb{R}^{d_M},\nu;\mathbb{R}^{d_M})$ norm, but instead attempts to construct $m$ to reduce errors for $q $ in $L^2_\nu$. Other related bounds for Monte Carlo for AS are investigated in \cite{ConstantineGleich2014}, and in a more general setting we note that subsampled eigenvectors are known to be good approximations of the true eigenvectors in settings such as matrix completion \cite{KarouidAspremont2009}. What matters most is not the norm of the error between two subspaces, but the \emph{subspace angles}, since the neural network training can easily account for rotations within a fixed subspace, while the errors above suggest a much worse state of affairs in such a case.

Additionally, in high dimensions, \eqref{general_stochastic_basis_problem} is typically solved matrix-free via randomized methods \cite{MartinssonTropp2020}, which introduces an additional stochastic error in the approximation of the dominant matrix action on a randomly sampled basis. Thus in practice there may be multiple sources of errors. In the following proposition we give a bound for how input reduced basis errors affect approximations of ridge function approximations of Lipschitz maps. 

\begin{proposition}{Bounds for the effects of input reduced basis errors on ridge function approximation.}\label{prop_basis_error}

Suppose there exists a ridge function approximation $q_{r}$ with orthonormal output reduced basis $\Phi_r$, and input reduced basis $V_{r}$ such that 
\begin{equation}
    \|q - \Phi_r \circ q_{r}\circ V_{r}^T\|_{L^2_\nu} < \epsilon_\text{ridge},
\end{equation}
and additionally orthonormal $\widetilde{\Phi}_r$, general $\widetilde{V}_{r}$ are approximations of the reduced bases $\Phi_r,V_{r}$ respectively, such that
\begin{align}
    &\|\Phi_r - \widetilde{\Phi}_r\|_{\ell^2(\mathbb{R}^{d_Q \times r})} < \epsilon_\text{output}\\
    &\|V^T_{r} - \widetilde{V}^T_{r}\|_{\ell^2(\mathbb{R}^{r\times d_M})} < \epsilon_\text{input},
\end{align}
and $q_{r}$ is Lipschitz with constant $L_{q_{r}}$, then
\begin{align}
    \|q - \widetilde{\Phi}_r \circ q_{r}\circ \widetilde{V}_{r}^T\|_{L^2_\nu} < \nonumber \\
     \epsilon_\text{ridge}(1 + \epsilon_\text{output}) + \epsilon_\text{output}\|q\|_{L^2_\nu} + L_{q_{r}} \epsilon_\text{input} \left[\sqrt{\text{tr}(\mathcal{C})} + \|m_0\|_2 \right]
\end{align}
\end{proposition}

See \ref{proof_prop_basis_error} for the proof. This bound shows that both the distribution $\nu$ and the worst case amplifications of the ridge function $q_{r_M}$ can amplify the input reduced basis errors. The effects of the distribution $\nu$ are decomposed into effects of the mean $m_0$ and the covariance $\mathcal{C}$; a larger mean and more variance in the distribution (as encapsulationed by $\text{tr}(\mathcal{C})$) can lead to larger excitations in erroneous modes of $\widehat{V}_{r_M}$. The worst-case propagation of these errors through the ridge function is captured by the Lipschitz constant $L_{q_{r_M}}$. The effects of the output basis error contribute mildly to the ridge function error but contribute another term involving the norm of $q$; if $\|q\|_{L^2_\nu}$ is very large, even small output basis errors can lead to large errors in absolute terms.





\section{Learning Reduced Basis Mappings via Adaptive ResNets} \label{section_control_flow}

In this section, we assume that we can approximate the map $m\mapsto q$ with acceptable accuracy by a reduced basis ridge function $\Phi_r \circ q_r \circ V_r^T \approx q$ with acceptably small $r$. The focus of this section is on how to construct a neural network $q_r(m_r) \approx f_{w,r}(m_r) = f(m_r,w):\mathbb{R}^{r}\times \mathbb{R}^{d_W} \rightarrow \mathbb{R}^{r}$ parametrized by weights vector $w \in \mathbb{R}^{d_W}$ that can detect an appropriate nonlinear representation for the reduced basis mapping in one constructive process. This can be combined with the constructive procedures for linear reduced basis approximation discussed in the last section to yield a unified constructive framework, mitigating issues regarding architectural hyperparameter tuning for breadth and depth. Given appropriate input and output reduced bases $\Phi_r,V_r$, our goal is two-fold.

\newcommand{\subscript}[2]{$#1 _ #2$}
\begin{enumerate}[label=\subscript{G}{{\arabic*}}.]
    \item (Universal approximation): We seek a surrogate approximation strategy such that for any $\epsilon >0$ there exists a neural network $f_{w,r}$, such that $\| q - \Phi_r \circ f_{w,r} \circ V^T_r \|_{L^2_\nu} < \epsilon$ . Since we have already handled the breadth via linear dimension reduction, the limiting sense in which the neural network approximates the map is via addition of depth to the neural network architecture, which we take to loosely be related to the nonlinearity of the map $m\mapsto q$.

    \item (Practical approximation) To have a reasonable algorithmic means of attempting to achieve high quality approximations, particularly given limited data. The sense in which the neural network is constructed ought to be connected to the sense in which the class of neural networks are universal approximators. 
\end{enumerate}

The reduced basis neural network can be thought of as a \emph{latent space} mapping between the coefficients of the $r$ dimensional input reduced basis and the $r$ dimensional output reduced basis. The learning problem (with respect to the true measure $\nu$) is formally expected risk minimization in $L^2(\mathbb{R}^{d_M},\nu;\mathbb{R}^{d_Q})$:
\begin{equation}
    \min_{w \in \mathbb{R}^{d_W}} \| q - \Phi_r \circ f_{w,r} \circ V_r^T \|_{L^2_\nu}^2.
\end{equation}
But given the restrictions of the learned approximation $f_{w,r}$ to only learn the representations in the $r$-dimensional coefficients of the reduced bases $\Phi_r,V_r^T$, one can equivalently conceive of the learning problem in the space $L^2(\mathbb{R}^r,\nu^{V_r};\mathbb{R}^r)$ where $\nu^{V_r}$ is the conditional measure for $\nu$ corresponding to the sigma-algebra generated by the input reduced basis $V_r$. The output is reduced by restricting the output learning problem to only the coefficients of $\Phi_r$. The input space is reduced by the conditional expectation taken with respect to the sigma-algebra generated by the $V_r$ basis, having the effect of marginalizing out contributions of the input uncertainty in the orthogonal complements of $V_r$. We denote the change of variable via conditional expectation for the truncated input parameter by $\mathbb{R}^r \ni m_r = V_r^m$ for $m \in \mathbb{R}^{d_M}$. The truncated input parameter is then sampled from the conditional probability measure: $m_r \sim \nu^{V_r}$ restricted to the coefficients of the input reduced basis.

The universal approximation goal ($G_1$) can be satisfied by any neural network architecture class that is proven to be a universal approximator of $L^2(\mathbb{R}^r;\mathbb{R}^{r})$ functions, of which there are many \cite{Cybenko1989,Hornik1991,LiLinShen2019,LinJegelka2018,LuPuWangEtAl2017}. This result can be extended to $L^2_\nu(\mathbb{R}^{r},\nu^{V_r};\mathbb{R}^r)$ given different assumptions on the map and or the measure. For example, if the density function associated with $\nu^{V_r}$ denoted by $\pi^{V_r}(m_r)dm_r = d\nu(m_r)$ is essentially bounded (by $C_{\pi^{V_r}}$), then
\begin{equation} \label{exp_risk_min}
    \int_{\mathbb{R}^{r}} \|f_{w,r}(m_r) \|_{\ell^2(\mathbb{R}^r)} \pi^{V_r}(m_r)dm_r \leq C_{\pi^{V_r}}\int_{\mathbb{R}^{r}} \|f_{w,r}(m_r) \|_{\ell^2(\mathbb{R}^r)}dm_r.
\end{equation}

A major issue for neural network based methods however is that $G_1$ does not necessarily guarantee $G_2$. Many such universal approximation theorems involve complexity estimates (depth, breadth) required to achieve a specific approximation accuracy.  Obtaining such approximations in practice is still out of reach for many reasons such as the non-convexity of the neural network training problem. Making things more difficult, the means in which neural networks are trained in practice bears almost no resemblance to the proof mechanisms typically used to establish their universal approximation properties. Neural networks are often proven to be universal approximators by demonstrating that a specified class of neural networks can be used to construct approximating units, which in turn are capable of arbitrarily well-approximating e.g., simple functions or polynomials; since these function classes are dense in $L^p$ functions, the universal approximation result then follows for $L^p$ spaces.  In practice neural networks are trained via the solution of a stochastic nonconvex empirical risk minimization problem (the discrete analog of \eqref{exp_risk_min} over sample data). While the theory posits that for a given neural network function class the global minimizer of the optimization problem may correspond to a suitable neural network approximation, there are no guarantees that one will obtain such an approximation in practice. 

Not yet taking into account the confounding effects of statistical sampling error, a fundamental issue in the gap between $G_1$ and $G_2$ is the complex relationship between the limiting sense of the approximation of a given neural network class (e.g., adding depth), and the associated training problem. While continually adding more representation complexity in a proof allows errors to be driven to zero, in many cases, modifying a neural network's structure (e.g., adding a layer to a fully-connected network) significantly distorts the neural network mapping and requires re-training. Adding too much depth as well is problematic; the empirical ``peaking phenomenon'' \cite{Hughes1968} demonstrates that often as neural networks become more expressive they get more difficult to train in practice, and ultimately lead to worse performance. This phenomenon is observed in the numerical experiments in Section \ref{section_numerical_experiments}.

To address the aforementioned issues, we propose the adaptive construction of ResNet approximations of the reduced basis mapping between $\mathbb{R}^r$ and $\mathbb{R}^r$. ResNet layers are nonlinear perturbations of the identity map; at layer $l+1$ the hidden neuron representation $z_{l+1}$ is defined by the recurrence relation:
\begin{equation} \label{discrete_resnet}
    z_{l+1} = z_l + w_{1l}\sigma_l(w_{0l}z_l + b_l),
\end{equation}
where $w_{1l},w_{0l}^T \in \mathbb{R}^{r\times k}$, $b_l \in \mathbb{R}^k$ are weight arrays, $\sigma_l$ is an element-wise nonlinear activation function, and $k$ is the dimension of the nonlinearity added at each layer. The initial layer takes the input data $m_r$, and the last layer output approximates the target output data $q_r \in \mathbb{R}^{r}$. The fundamental observation is that the nonlinear approximation power of ResNets is in their depth, and of essential importance to this work, the architecture can be marginally perturbed, since adding a new layer with $\|w_{1l}\sigma_l(w_{0l}z_l + b_l)\|_2 \leq \epsilon$ perturbs $z_l$ by less than $\epsilon$. This is in contrast to other popular neural network building blocks such as dense feedforward; in this case the addition of a new neural network layer nonlinearly distorts the latent space mapping. For this reason ResNets can be trained and constructed iteratively in one procedure, while fully-connected feedforward networks cannot. ResNets are capable of producing high quality approximations given few weight parameters, as noted in \cite{LinJegelka2018}, the identity map adds significant expressive capabilities to the surrogate and requires no weights to do so. In what follows we are able to build on recent approximation theory \cite{LiLinShen2019} that connects ResNets with sequentially minimizing control flow ODE systems, and use this theory to prove that reduced basis ResNet architectures are universal approximators of $L^2_\nu$ functions, which satisfies our first goal ($G_1$). This conception of ResNets as discrete sequential minimizing flows motivates the practical adaptive ResNet construction algorithm that we present, which satisfies our second goal ($G_2$). Moreover, we discuss practical issues such as the effects of statistical sampling errors incurred in the neural network training (empirical risk minimization) problem.

\subsection{ResNets and Control Flows}

A residual neural network can be thought of as an explicit Euler approximation of a control flow ODE (often referred to as neural ODEs) \cite{ChenRubanovaBettencourtEtAl2018,LiLinShen2019,RuthottoHaber2020}. The control flow evolves the input data  $m_r \in \mathbb{R}^{r}$ to the corresponding target data $q_r(m_r) \in \mathbb{R}^r$ by finding a parametrization of the ODE that minimizes a specified objective function, in our case the square of the $L^2_\nu$ norm. This can thus be stated as an ODE optimal control problem with control variable $w(t)$:

\begin{subequations} \label{generic_control_ode}
\begin{align}
    \min_{w(t)} &\|z(T) - q\|_{L^2_\nu}^2\\ 
    \frac{dz}{dt}(t) &= \phi(z(t);w(t)) \\
    z(0) &= m_r = V_r^Tm.
\end{align}
\end{subequations}
The right hand side that is the continuous time analog of \eqref{discrete_resnet} is
\begin{equation} \label{node_rhs}
    \phi(z,t) = w_1(t) \sigma(t)(w_0(t) z + b(t))
\end{equation}
where for each time $t$, $w_1(t) \in \mathbb{R}^{r\times k},w_0(t) \in\mathbb{R}^{k\times r}, b(t)\in \mathbb{R}^k$, where $k\leq r$ is a architecture parameter,  and $\sigma(t)$ is a nonlinear function that is applied elementwise. In \cite{LiLinShen2019}, the authors show that a sufficient condition for the class of functional approximations by generating flows \eqref{generic_control_ode} to be universal approximators of continuous functions on compact sets is that functional representation of the right hand side for the ODE satisfies conditions that are highly relevant to ResNets. First the class of functions for the right hand side is closed under affine operations, and second is that the closure of the class contains a ``well function'' that can be used to modify certain regions of $\mathbb{R}^r$ while leaving others untouched. For example, ReLU is a well function because it maps the entire left half-plane to zero. Many other popular activation functions in machine learning are well functions. Using the well function property, the universal approximation proofs in \cite{LiLinShen2019} construct sequential control flows that evolve local regions of latent space representation to continue moving trajectories of the initial condition data to their corresponding targets in the output. The locality of the action of the right-hand side of the control flow ODE is key to this proof construction; it additionally suggests choosing the layer rank hyperparameter $k \ll r$ has benefits beyond reducing the complexity of the weights. Indeed for a given ResNet layer mapping \eqref{discrete_resnet}, the latent representation update $z_l \mapsto z_{l+1}$ modifies the latent representation in only the span of $w_{1l}$, in a mapping that is only sensitive to the latent representation in the span of $w_{0l}^T$. The nonlinear activation function has the effect of further concentrating this update as well adding the critical nonlinearity. 

In a limiting fashion this construction proves that there exists a finite time ($T>0$) control flow that can get arbitrarily close to approximating any continuous function on a compact set $K \subset \subset \mathbb{R}^r$. We extend this to our settings and state a representation error bound for reduced basis ResNet approximations of $L^2_\nu$ parametric mappings.

\begin{theorem}{Representation Error Bound for Reduced Basis ResNet} \label{projected_resnet_representation_theorem}

Given a parametric mapping $q \in L^2(\mathbb{R}^{d_M}, \nu; \mathbb{R}^{d_Q})$ that can be approximated by restriction to rank $r<\min\{d_M,d_Q\}$ reduced bases of the inputs with truncation error given by:
\begin{equation}
\|q - \Phi_r  \circ q_r \circ V_r^T\|_{L^2_\nu} \leq  \zeta_r < \infty
\end{equation}
where $q_r:\mathbb{R}^r \rightarrow \mathbb{R}^r$ is a ridge function that approximates $q$ between the reduced basis for the inputs $V_r$ and the orthonormal reduced basis for the output $\Phi_r$. Assuming the density $\pi(m)dm = d\nu(m)$ corresponding to $\nu$ is essentially bounded, then for any compact set $K \subset \subset \mathbb{R}^{d_M}$ there exists an input-output projected ResNet $f_{w,r}(V_r^Tm)$ such that 
\begin{equation}
\|q - \Phi_r  \circ f_{w,r}\circ V_r^T\|_{L^2(K,\nu;\mathbb{R}^{d_Q})} \leq  2\zeta_r 
\end{equation}
and the depth of the ResNet is $O\left(\frac{Te^T|K|}{\zeta_r}\right)$, where $T$ is the time horizon for the associated control flow mapping that is the continuous time analog of the ResNet.
\end{theorem}

See \ref{representation_error_appendix} for proof. Additionally noted in the appendix, are conditions under which the depth complexity can be reduced to $O\left(\frac{T|K|}{\zeta_r}\right)$, which is related to details of the construction of the control flow being approximated via explicit Euler. 

Since the truncation errors can be made arbitrarily small by the choice of rank for the basis, this result establishes that high dimensional parametric mappings restricted to compact sets can be approximated arbitrarily well by reduced basis ResNets, and establishes an approximation bound that depends on the truncation error as a result of the projections, as well as the complexity in the ResNet approximation of the reduced ridge function. In particular when the high dimensional map can be well-approximated for $ r \ll \min\{d_M,d_Q\}$ the weight complexity of the ResNet can be reduced significantly compared to overparametrized networks. As the result states, the ResNet depth is inversely proportional to the desired accuracy, and proportional to the size of the desired region where the map is to be approximated $|K|$, as well as a complexity upper bound for the approximation of the restricted nonlinear map via a control flow with time horizon $T$. Inner regular measures (such as Gaussians) can be approximated arbitrarily well on compact sets; that is, for any $\epsilon >0$, there exists a compact set $K_\epsilon$ such that the measure of $K_\epsilon^c$ is bounded by $\epsilon$. Other distributions such as uniform distributions in finite dimensions trivially have compact support. This makes the result sufficiently general, however extending this result to all of $\mathbb{R}^{d_M}$ is tricky since the depth complexity for the ResNet is a function of $|K_\epsilon|$. More sophisticated bounds could make use of results in concentration of measure in order to mitigate the dependence on $|K|$. 


The result hinges on the existence of a finite time horizon approximating control flow for the target function. Unfortunately the result does not say anything generally about how large $T$ is as a function of the complexity of the target function $q_r$. In the case of 1D-monotonic functions a bound is given in \cite{LiLinShen2019} in terms of the regularity of the target function; when the total variation of the Jacobian of the map is smaller, less time is needed to move the input data space to the output data space. In future work, we expect that these approximation rates can be extended to high dimensional non-monotonic functions; one would expect that smoother functions are easier to approximate. 


\subsection{Effects of Statistical Sampling Error and Training}

In practice one does not have direct access to the statistical measure $\nu$, and instead integration with respect to $\nu$ is approximated empirically by Monte Carlo. This leads to empirical risk minimization over sample data $\{(m_i,q(m_i)|m_i \sim\nu\}_{i=1}^N$,
\begin{equation}
  \min_w\frac{1}{N}\sum_{i=1}^N \|q(m_i) - \Phi_rf_{w,r}(V_r^Tm_i)\|_{\ell^2(\mathbb{R}^{d_Q})}^2
\end{equation}
and additionally the Monte Carlo approximations of the reduced bases (discussed in Section \ref{effects_of_subsampling_subsection}). At this point it is useful to bound the \emph{generalization gap}, i.e., the difference between the empirical risk and expected risk obtained from the sampling of the reduced bases, and the solution of the empirical risk minimization problem. Such a bound requires delicate treatment due to the complex correlation between the statistically estimated reduced bases and solution to the empirical risk minimization problem, and is outside of the scope of this work. Typically such a bound would establish a sample complexity required for both the estimation of the reduced bases and the empirical risk function in order to achieve a certain generalization accuracy \emph{in expectation}. In practice, however, it is worth noting that such a bound may be conservative, because the empirically trained neural network $f_{w,r}$ can make up for errors in the reduced basis approximations during training, while combined bounds may penalize the neural network for doing so. Achieving such a bound may be very complicated for deep neural networks due to the difficulity in unifying the associated approximation power of the function class and the effects of statistical sampling error. The interested reader can refer to e.g., \cite{CuckerSmale2002,deHoopKovachkiNelsenEtAl2021} for examples of such analyses, for e.g., linear regression.

In general, such bounds establish that while the reduced basis methods posit the existence of compressed representations of $q \in L^2_\nu$, statistical sampling errors may be a fundamental limitation to achieving such approximations in practice. Specifically such bounds are useful to demonstrate the trade-off between the sample complexity and the variance of what is being estimated statistically; when there is more variance, more samples may be needed to obtain faithful approximations. If such a bound exists, it may posit that with a sufficient amount of data to well approximate the reduced bases and empirical risk function, one could in principle attain a sufficiently generalizable surrogate. This however can be misleading for deep neural networks due to the additional errors incurred in the neural network training, i.e., the difference between the global minimizer and the local minimizer achieved in practice. Such bounds are hard to approximate generally, and lie outside the scope of this work; but generally a gap in the optimization objective at an obtained local minimizer versus the objective function and (a) global minimizer represents another fundamental limitation for practical machine learning. 

In summary our approach attempts to improve performance given limited samples by alleviating overparametrization in the limited data regime and limiting the need for architectural hyperparameter tuning. The goal is that by reducing the number of inferred weights, the ill-posedness inherent in the high-dimensional, limited data training problem will be ameliorated, and the weights that do get inferred are meaningful to the underlying map because they learn the mapping between only informed reduced bases of the inputs and outputs. The performance benefits of this approach are demonstrated in numerical experiments in Section \ref{section_numerical_experiments}.

\section{Overview of Framework and Algorithm} \label{section_algorithm_overview}

In this section we overview the approach at a high level and discuss practical algorithmic considerations. The algorithm proceeds in two main stages: first linear dimension reduction via restriction to $\Phi_r,V_r$, then adaptive construction of the nonlinear representation of the mapping between the two bases. 

\subsection{Step 1: Linear Dimension Reduction}
The first step is of critical importance, since if the bases $\Phi_r,V_r$ are not chosen wisely, critical information for the mapping $m \mapsto q$ will be lost by the effects of the basis truncations. This step involves two decisions, first: what are appropriate $\Phi_r,V_r$, and second what is an appropriate basis dimension $r$. In Section \ref{reduced_basis_surrogate_section} we discuss error bounds for POD, KLE, and AS that can be used to guide the choice of the basis dimension for these three bases. In the case of POD, appropriate choice of $r$ is dictated by the decay of the eigenvalues of $\mathbb{E}_\nu[qq^T]$ which are empirically estimated in the computation of the POD basis. Additionally one can non-intrusively estimate the related error $\|(I_{d_Q} - \Phi_r\Phi_r^T)q\|_{L^2_\nu}$ via sample average approximation in order to check if a given $r$ is appropriate for a desired error tolerance (up to the effects of statistical estimation error). The story is not quite the same for the choice of $r$ for KLE and AS. In the case of KLE, the error estimations involve both the eigenvalue decay of the covariance $\mathcal{C}$ as well as the Lipschitz constant for $q$. Estimating an appropriate choice of KLE basis rank using this bound thus would involve an empirical estimation of the Lipschitz constant for the map; and since this bound is conservative it may suggest a much larger basis rank than what is actually required in practice. For active subspace the error estimations are related to the generalized eigenvalue problem for the construction of the AS basis \eqref{active_subspace}, so like POD the appropriate choice of rank for AS can be estimated directly from computations used in the construction of the basis (again up to the effects of statistical estimation error). Unlike POD, however for both AS and KLE, the additional error estimation $\| q - q\circ V_rV_r^T\|_{L^2_\nu}$ cannot be estimated non-intrusively. Indeed doing so requires additional queries of the expensive map $m\mapsto q$, which may be better spent on the computation of additional training data. In all cases, one should take into account statistical estimation errors when estimating appropriate $r$ for any basis that requires Monte Carlo approximation. The linear dimension reduction ideas in Section \ref{reduced_basis_surrogate_section} are sufficiently general to be extended to other settings; we discuss POD, KLE and AS because of their connection to $L^2_\nu$ functions. In our construction we assumed $q$ to be mean zero. This assumption can be handled at this step by using a sample average mean of the outputs $q$ as a last layer bias, so that the neural network in the next step just focuses on learning the deviations from the sample average mean via the reduced basis ResNet.

\subsection{Step 2: Adaptive Nonlinear Learning}

In this step the nonlinear ResNet approximation of the mapping $\mathbb{R}^r \ni V_r^Tm \mapsto \Phi_r^T q(m)\in\mathbb{R}^r$ is learned. Step one decides an appropriate choice of $r$. Left to be decided in this step are the choice of activation function $\sigma$, the layer rank $k$, and the depth of the ResNet approximation. The choice of activation function should be guided by the well function property discussed in \cite{LiLinShen2019}; e.g., popular activation functions ReLU, softplus, sigmoid, etc., satisfy this property. As discussed in Section \ref{section_control_flow} we suggest taking $k \ll r$ because learning compressed (low-dimensional) modifications of the latent space representation is fundamental to the universal approximation property of control flows; additionally it has the additional benefit of keeping the weight dimension small, making the associated sequential optimization problems simpler. For the sake of simplicity, choosing the activation function and layer rank to be fixed for all layers, the final architectural hyperparameter, the depth, is found algorithmically. 

The reduced basis ResNet is constructed iteratively, layer by layer, with a few decisions required by the user to decide how the training is done, and when to stop. We suggest using stochastic optimizers that are typical to neural network training (e.g., Adam or stochastic gradient descent), for the layer-wise training. In order to avoid overfitting early in the algorithm, we suggest deliberately under-training at each iteration of the algorithm. A major decision for the approach is what is trained at each iteration? In the truly sequential control flow problem, one would train only the most recently added layer, leaving all of the previous layer weights fixed. Training all layers in each iteration, while adaptively adding layers has the benefit of a larger configuration space for the optimizer to explore. The last major decision for the algorithm is when to terminate: we suggest monitoring a discrepancy between the validation accuracy and training accuracy, and to use this as a heuristic estimate of over-fitting. In the sequential training version of the algorithm, once over-fitting is detected, the previous layers can be frozen, and the algorithm can continue to see if adding additional layers (trained one-by-one) improves the accuracy; this has been a useful strategy for us in practice. In the case that all layers are trained at once, the over-fitting heuristic can be used to terminate the construction of the ResNet. After the neural network is constructed we suggest doing one additional end-to-end training in order to see if the approximation can be improved. In our experiments, we use the Adam optimizer for the constructive phase of the algorithm and a stochastic Newton-based neural network training algorithm with small step sizes \cite{OLearyRoseberryAlgerGhattas2019,OLearyRoseberryAlgerGhattas2020}, which empirically helps improve generalization accuracy when close to a local minimum. We give an overview of all steps of our method in Algorithm \ref{alg:rb_resnet_training}.

\begin{algorithm}[H]
\begin{algorithmic}[1]
\STATE Given reduced bases for inputs and outputs: $V_r,\Phi_r$, and sample data $\{(m_i,q_i)|m_i\sim\nu\}_{i=1}^{N}$
\STATE $z_0(m) = V_r^Tm$ (truncated representation of input data)
\STATE $b_Q = \frac{1}{N}\sum_{i=1}^N q_i$
\STATE $l=0$
\WHILE{Validation accuracy is increasing}
  \STATE $l = l + 1$
    \STATE Partially train Projected ResNet $f_{w,r}(m) = \Phi_r z_{l}(m;w) + b_Q$\\
     via empirical risk minimization for a small number of epochs \\
      (e.g., using Adam, with fixed batch, step length),\\
    with added ResNet layers, i.e.,\\
     $z_{l}(m;w) = z_{l-1}(m) + w_{l-1,1}\sigma(w_{l-1,0}z_k(m) + b_{l-1})$ as in \eqref{discrete_resnet}
\ENDWHILE
\STATE One final end-to-end training (using stochastic Newton optimizer)
\RETURN Projected ResNet $\Phi_r \circ f_{w,r} \circ V_r^T + b_Q \approx q$
\end{algorithmic}
\caption{Reduced Basis ResNet Construction}
\label{alg:rb_resnet_training}
\end{algorithm}

\section{Numerical Experiments} \label{section_numerical_experiments}

We investigate the performance of adaptive reduced basis ResNets on problems stemming from parametric PDE maps, and an aerodynamic shape optimization problem. We compare our proposed adaptively trained reduced basis ResNets to various other architecture strategies such as end-to-end trained reduced basis ResNets, reduced basis dense networks, and encoder-decoder networks where pre-computed reduced bases are not used. In the case of end-to-end trained reduced basis ResNets, the same architectures are used as in the adaptively trained analog, however all layers are trained at once, instead of using the adaptive construction and training delineated in Section \ref{section_algorithm_overview}. The reduced basis dense networks use fully-connected dense networks to learn the mapping between the same reduced bases used in the reduced basis ResNets. In the case of the encoder-decoder networks, no pre-computed reduced bases are used in the construction of the network, and instead the network learns input and output subspaces as the first and last layers of the networks; thus in this case the weight dimensions depend directly on $d_M,d_Q$. For the encoder-decoder networks we use both ResNet representations as well as fully-connected dense layers; descriptions and names for these networks are given in each example. For each set of results we demonstrate the performance of the different architectures in relation to the number of data samples seen during training, as well as in relation to differing depths and breadths.

For all numerical results we train the networks using the Adam optimizer, with step length$\alpha = 10^{-3}$. We use Glorot uniform initialization in TensorFlow \cite{AbadiBarhanChenEtAl2016}, and no regularization for the empirical risk minimization. For all networks except for the dense reduced basis networks, we found using small batch sizes of $2$ improved the training, and all of these networks were trained for $50$ total epochs (in the case of the adaptive training, these epochs are split among individual trainings). In the case of the reduced basis dense networks, larger batch sizes with more optimization epochs were required than the other methods. For the parametric PDE problems we additionally employed a Newton optimizer for a final training for all networks, which led to improved performance, as discussed in \ref{appendix:two_step_opt}. We use softplus activation functions for all networks, which were found to work better than other activation functions. Additionally we set the layer rank $k=4$ for all problems, satisfying $k \ll r$ as was discussed in previous sections. For simplicity we consider the effects of adaptive depth-wise training with pre-defined fixed terminating depths. This algorithm can be modified to instead implement  termination criteria, as is discussed in Section \ref{section_algorithm_overview} 

In summary the numerical results demonstrate that the reduced basis ResNets are able to outperform the other methods, in particular given few training data, and when $d_M$ and $d_Q$ are very large. These methods kept the weight dimensions $d_W$ very small, and the iterative learning of compact perturbations in the reduced basis latent space led to a robust algorithm for adding nonlinearity to the surrogate approximation.

\subsection{Learning Parametric PDE Maps}

We consider parametric PDE problems, where $m$ represent a random coefficient field that parametrizes the PDE, and $q$ represents an output function of the PDE state. We consider two cases: $q$ represents the full state, and $q$ represents a pointwise observation operator applied to the state, as is common in inverse problems. In both cases we use centered Gaussian distributions $\nu = \mathcal{N}(0,\mathcal{C})$ for the parameter field. Here $\mathcal{C} = \mathcal{A}^{-2}$ is a trace-class Mat\'{e}rn covariance, where the elliptic operator $\mathcal{A}$ is defined as
\begin{equation}
  \mathcal{A} = (\delta I - \gamma \Delta)
\end{equation}
in the physical domain $\Omega$, and solved with homogeneous Neumann boundary conditions. The correlation structure for the distribution $\nu$ depends on the choices of $\delta,\gamma$; spatial correlation length is dictated by the ratio $\sqrt{\frac{\delta}{\gamma}}$, and for fixed correlation length the marginal variance is reduced by making $\gamma$ and $\delta$ larger. For these problems we use a sample approximation of the $L_\nu^2$ accuracy to measure the performance of a network, which we define as 

\begin{equation}
  L_\nu^2 \text{ accuracy } = 1 - \sqrt{\mathbb{E}_\nu\left[\frac{\|q - f\|^2_{\ell^{2}(\mathbb{R}^{d_Q})}}{\|q\|^2_{\ell^{2}(\mathbb{R}^{d_Q})}} \right]}.
\end{equation}

In order to emphasize the sensitivity to the training data, we rerun each test for $5$ different instances of training data and report median plus or minus one fifth credibility intervals. For the active subspace projector computation we use $128$ samples; this requires additional offline computations as discussed in Section \ref{reduced_basis_surrogate_section}. 

\subsubsection{Poisson Parameter-to-State Map}
We first consider a Poisson problem in a unit square physical domain $\Omega = (0,1)^2$, where the parameter $m$ represents a log-normal diffusion coefficient. This is a common test problem in parametric PDE inverse problems. This example is taken directly from a tutorial for \texttt{hIPPYlib} \cite{VillaPetraGhattas20}.

    \begin{align*}
    -\nabla \cdot(e^{m}\nabla u) = 0 \text{ in } \Omega \nonumber\\
    u = 1 \text{ on } \Gamma_\text{top} \nonumber \\
    e^m\nabla u \cdot n = 0 \text{ on } \Gamma_\text{sides} \nonumber\\
    u = 0 \text{ on } \Gamma_\text{bottom} 
    \end{align*}
\begin{figure}[H]
\center
\includegraphics[width = \textwidth]{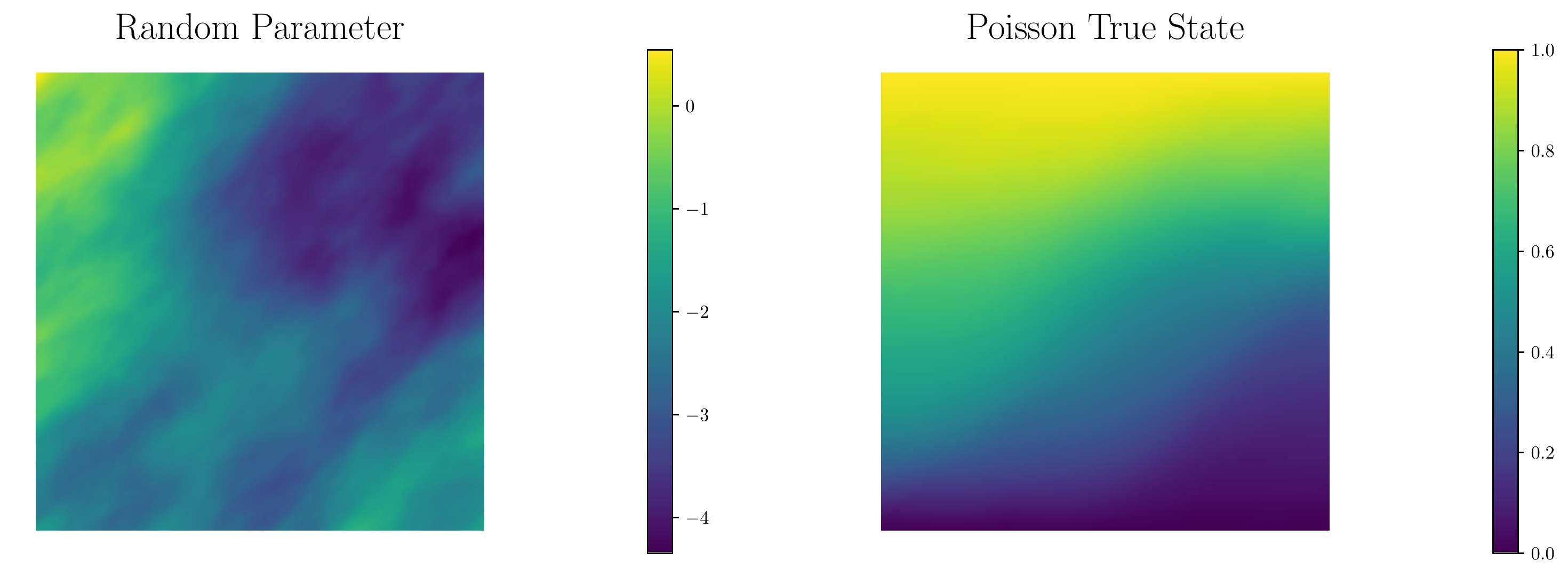}
\caption{Left: sample of parameter $m$, right: corresponding state solution.}
\label{fig:rdiff_state}
\end{figure}

We discretize $\Omega$ with a uniform $64 \times 64$ mesh, using linear finite elements for both $m$ and $u$. We use linear and quadratic basis functions for $m$ and $u$ respectively; the resulting input and output dimensions are thus $d_M = 4,225$ and $d_Q = 16,641$. The PDE state is driven by the boundary conditions: Dirichlet on $\Gamma_\text{top}=\partial\Omega\cap \{\mathbf{x} \in \Omega | x_2 = 1\}$) and $\Gamma_\text{bottom}=\partial\Omega \cap \{\mathbf{x} \in \Omega |x_2 = 0\}$. Homogeneous Neumann boundary conditions are used on $\Gamma_\text{sides} = \partial\Omega \setminus (\Gamma_\text{top} \cup \Gamma_\text{bottom})$. This problem demonstrates a key issue that makes parametric PDE learning hard at scale: discretizations of fields can get very large (e.g., millions or billions), and learning a mapping between two high-dimensional fields given limited training data is severely ill-posed. This problem is an excellent demonstration of the benefits of the reduced basis neural network methodology, since for this coercive elliptic problem the mapping $m \mapsto q$ is highly compressible, as our results demonstrate. 

\begin{figure}[H] 
\begin{subfigure}{0.5\textwidth}
\includegraphics[width = \textwidth]{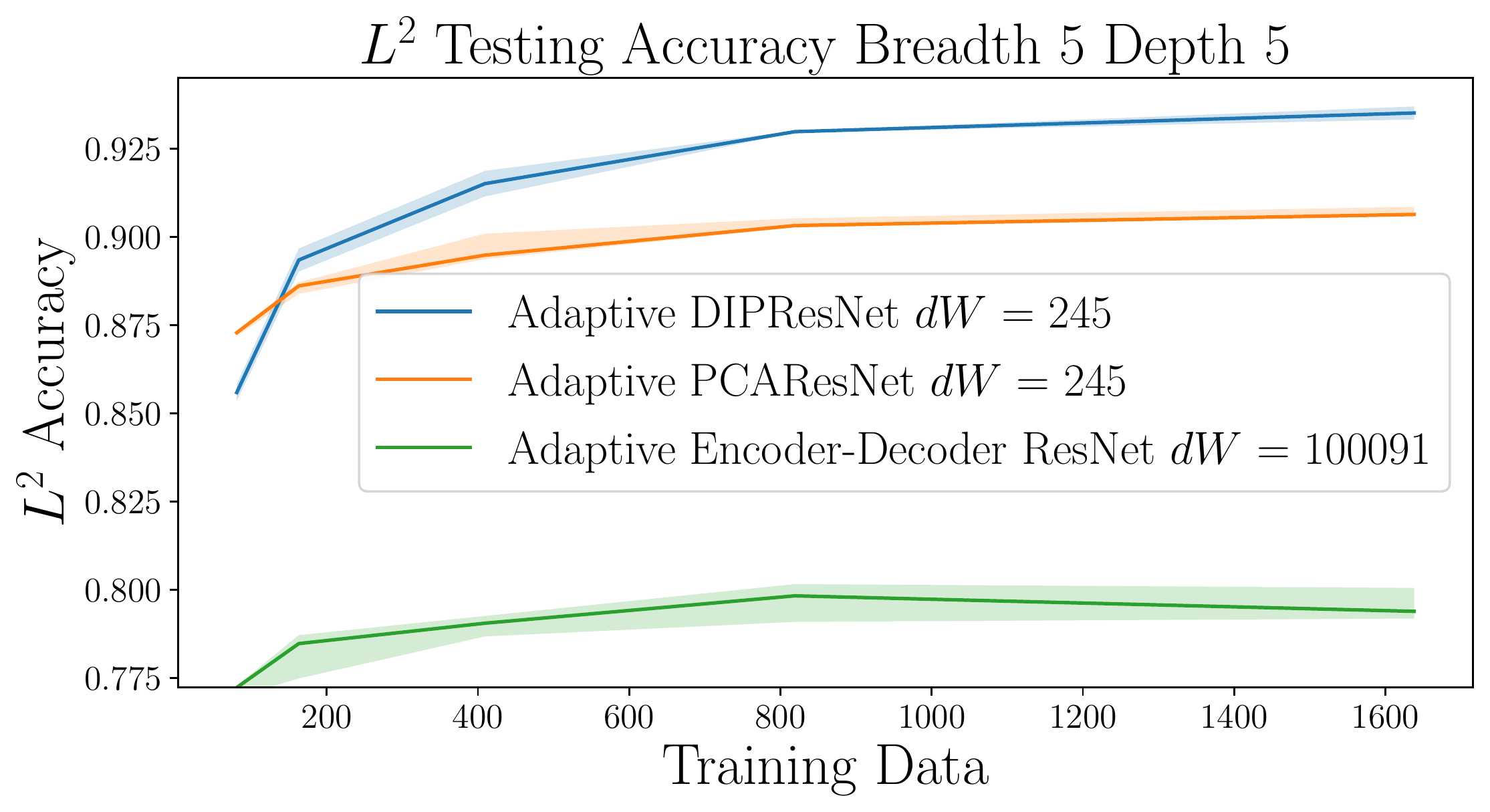}
\end{subfigure}%
\begin{subfigure}{0.5\textwidth}
\includegraphics[width = \textwidth]{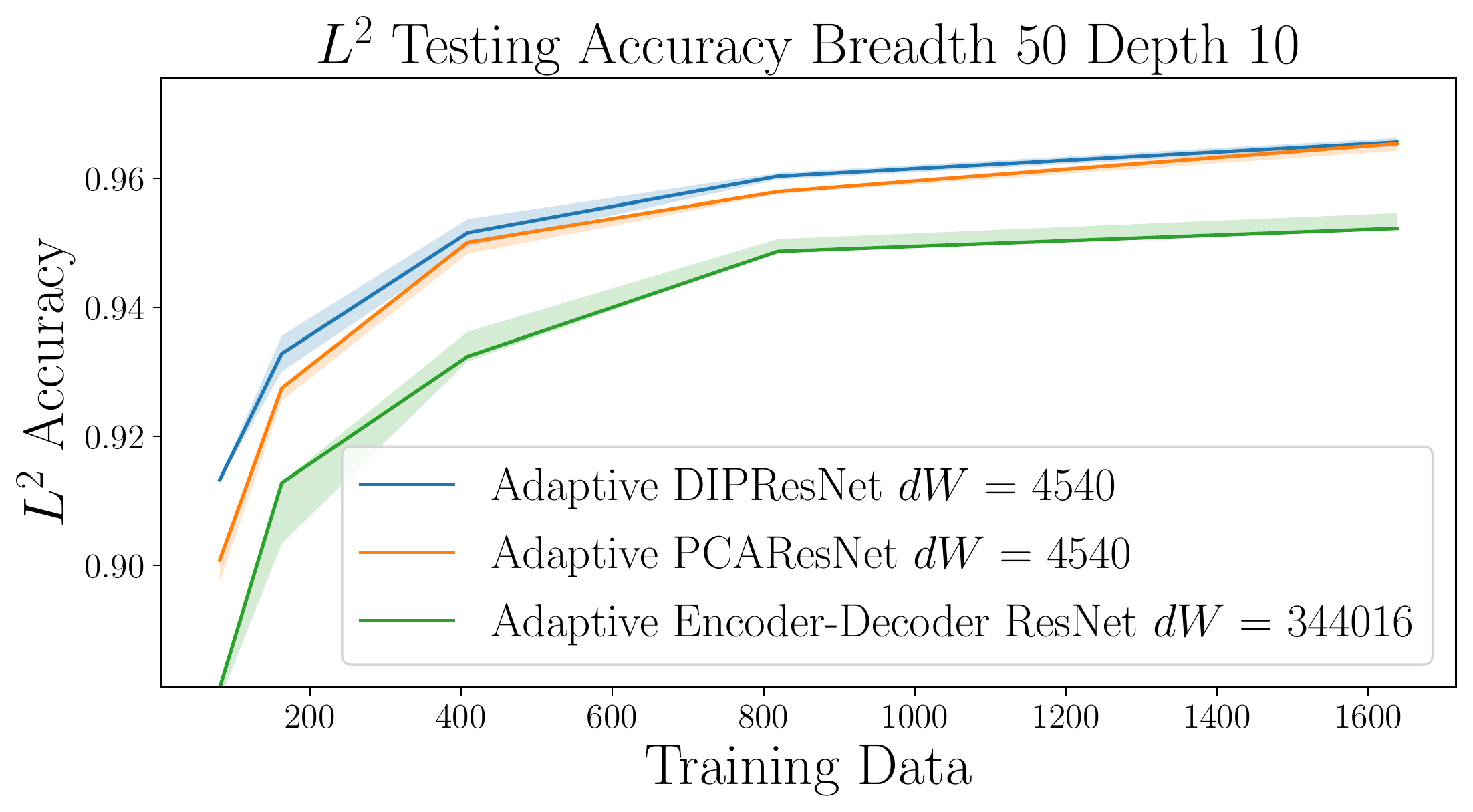}
\end{subfigure}
\caption{Comparison of adaptively trained ResNets. Two reduced basis strategies (DIPResNet which uses AS input basis, and PCAResNet which uses KLE input basis) are compared to an encoder-decoder strategy where the input and output bases are learned from a random initial guess via training. In all cases the DIPResNet network wins. In the right plot, the breadth and depth of the network are increased dramatically, and in this case the PCAResNet performs more similarly to the DIPResNet.}
\label{poisson_resnet_comps}
\end{figure}

For this problem we test a number of neural network architectural strategies: we use AS-to-POD reduced basis architectures \cite{OLearyRoseberryVillaChenEtAl2022} which we refer to as DIPResNet and DIPNet in the fully-connected dense case, and additionally the KLE-to-POD analog \cite{BhattacharyaHosseiniKovachki2020} which we analogously refer to as PCAResNet and PCANet. Numerical results demonstrate that the adaptive training of the ResNet improved accuracy over an end-to-end training, and additionally that the ResNet architecture led to a better, more compact representation than the fully-connected dense representation for the reduced basis mapping (DIPNet and PCANet). Due to the high-dimensional nature of the inputs and outputs for this problem we were not able to reliably train fully-connected dense networks without the use of reduced basis dimension reduction, and therefore these networks were omitted. Indeed, the inability of black-box neural network representations to approximate the input-output map motivates the developments presented here. For an encoder-decoder network that does not employ a reduced basis computed offline, we train an adaptive ResNet where the input and output layers are also trained from a random initial guess. The architecture is the same as DIPResNet and PCAResNet, but the reduced bases are found in the neural network training instead of via a priori computation. We refer to this network as an ``Adaptive Encoder-Decoder'' in Figure \ref{poisson_resnet_comps}.

For each run we compare the effects of breadth, depth, and training data on the generalization accuracy of the surrogate. We use a total of $512$ testing data. We begin by noting that the fully-connected dense DIPNet and PCANet were hard to train for this problem. In our experience the DIPNet dense network is somewhat sensitive to hyperparameters of the optimizer (which optimizer, step size, etc.). In \cite{OLearyRoseberryVillaChenEtAl2022} an inexact Newton CG optimizer with a line search and large batch size was used to obtain the best performance. Here we use Adam with batch size $32$ and the results do not compare favorably with the analogous ResNet. 

\begin{figure}[H] 
\begin{subfigure}{0.5\textwidth}
\includegraphics[width = \textwidth]{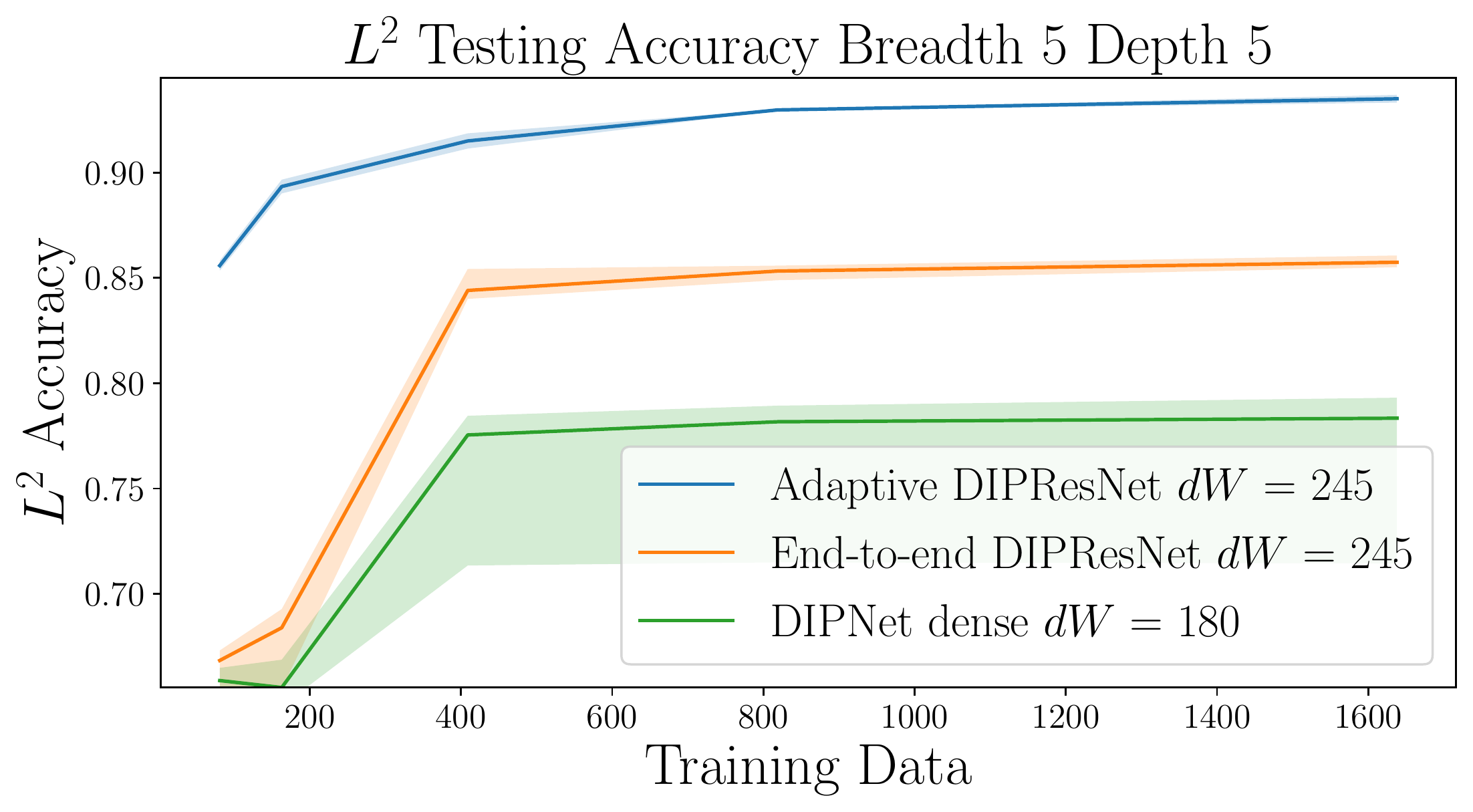}
\end{subfigure}%
\begin{subfigure}{0.5\textwidth}
\includegraphics[width = \textwidth]{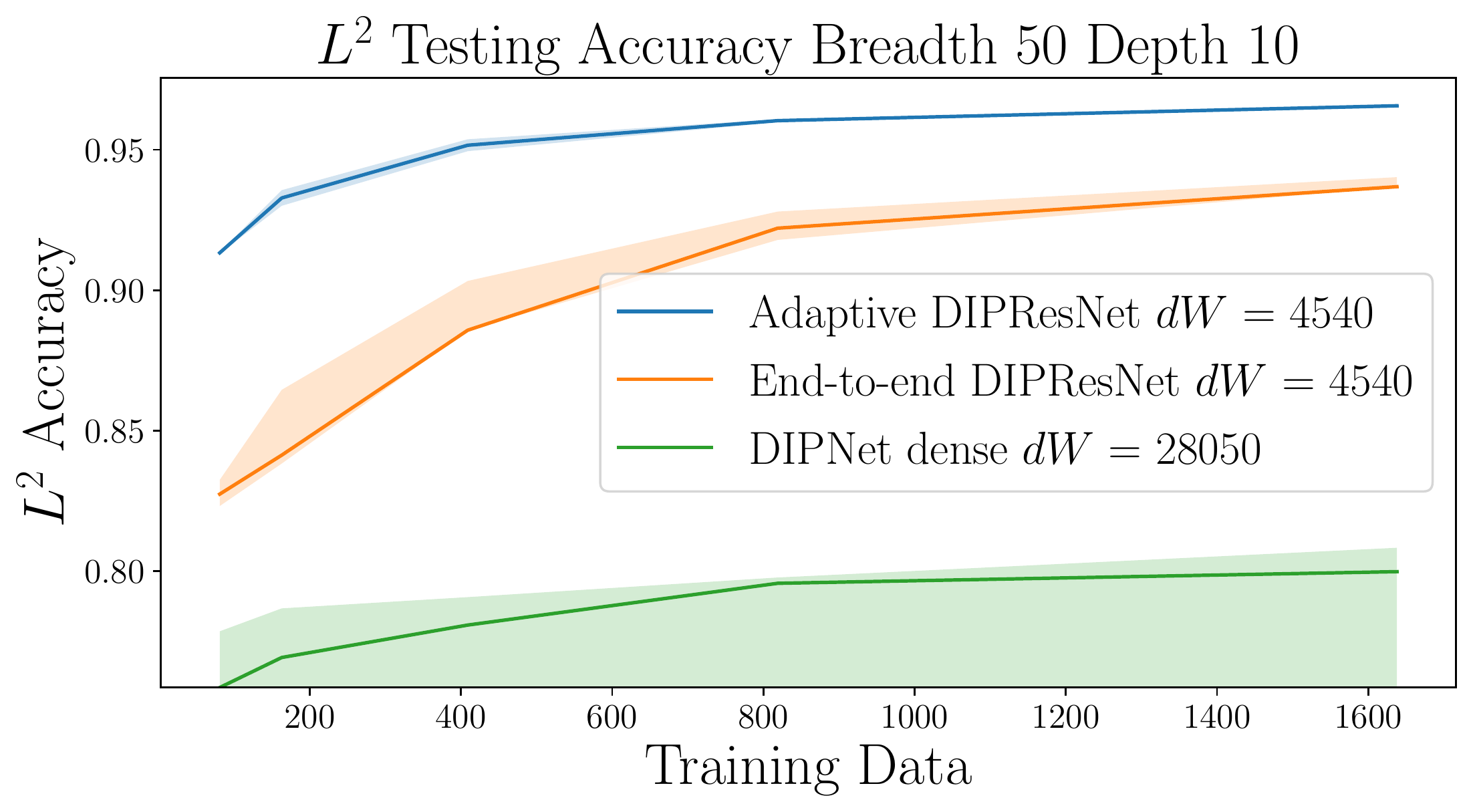}
\end{subfigure}
\caption{Comparison of different AS reduced basis networks. In all cases the adaptive DIPResNet outperforms the other networks, the identical end-to-end network trained in one shot yielded poorer performance than the adaptively trained network. The DIPNet architecture that uses a fully-connected dense representation of the reduced basis coefficient mapping performed worse than both DIPResNets. Similar results hold for PCA based architectures.}
\label{poisson_dipnet_comps}
\end{figure}

Next we demonstrate how the reduced basis neural network strategy compares to an identical ResNet structure, but where the bases are taken to be neural network weights, and are learned during the neural network training problem. Note that the weight dimensions for this network increase by several orders of magnitude.

Figure \ref{poisson_resnet_comps} demonstrates that the map is remarkably compressible: indeed approximation using a rank of only $5$ of the $4,225$ to $16,641$ dimensional map was able to achieve $>92.5\%$ accuracy (with the DIPResNet). This result required adaptive training. The nonlinear encoder-decoder network was not able to capture this low dimensional map. As depth and breadth were added to the architecture, however the encoder-decoder structure began to improve, and the differences between the AS basis and the KLE basis became much smaller. Since the linearized PDE map shares similar eigenvalue structure to the covariance $\mathcal{C}$, we would expect AS and KLE to perform very similarly asymptotically. 

What is remarkable about this approach is the scalability of the ResNet architectures. Indeed, with well chosen reduced bases, models with weight dimensions in the hundreds to thousands can give reasonable parametric approximations of these high-dimensional maps. For this case the encoder-decoder strategy eventually performs more similar to the DIPResNet and PCAResNet, but it requires $O(10^5)$ weights to catch up to the mere thousands the ResNet needs. When $d_Q,d_M$ are in the millions of dimensions the encoder-decoder ResNet strategy is no longer viable. Indeed for $d_Q,d_M$ in the tens of thousands, we were no longer able to train fully-connected encoder-decoders reliably. 

\subsubsection{Helmholtz Parameter-to-Observable Map}

We next consider a parametric map arising from a Helmholtz problem, which represents acoustic wave scattering in a heterogeneous medium. The construction of accurate and inexpensive surrogates from limited training data would make tractable the solution of high dimensional Bayesian inverse problems and Bayesian optimal experimental design problems, in particular those governed by wave propagation \cite{Bui-ThanhGhattasMartinEtAl13a, Bui-ThanhGhattas14, Bui-ThanhGhattas12a, Bui-ThanhGhattas12, AlexanderianGloorGhattas16, WuOLearyRoseberryChenEtAl22}. In this problem the parameter $m$ represents an uncertain medium, and the quantity of interest $q$ is the pointwise observation of the corresponding total wave field $u$ at 200 points $\mathbf{x}_i \in \Omega = (0,3)^2$, which is the physical domain.

\begin{align*}
-\Delta u - e^{2m}k^2 u = f \nonumber \\\text{ in } \Omega = (0,3)^2 \nonumber\\
\text{PML on } \partial \Omega \setminus \Gamma_\text{top}\\
 u  = 0 \text{ on } \Gamma_\text{top} \nonumber\\
q(m) = [u(\mathbf{x}_i,m)]
\end{align*}
\begin{figure}[H]
\center
\includegraphics[width = \textwidth]{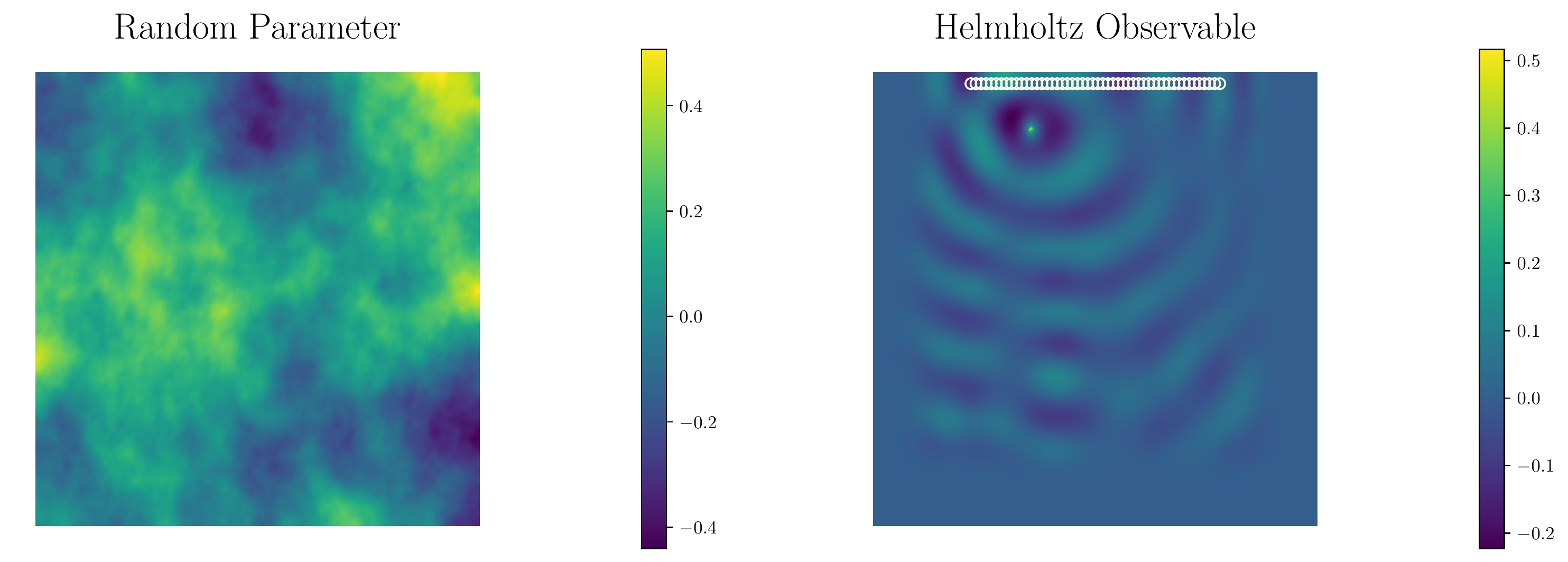}
\caption{Left: a random parameter sample $m$, right: corresponding wave field and observation locations.}
\label{fig:rdiff_state}
\end{figure}

For this problem the wavenumber is $4.55$. The right hand side $f$ is a unit point source located at $\mathbf{x} = (0.77,2.5)$. The perfectly matched layer (PML) boundary condition approximates domain truncation for a semi-infinite domain (waves are allowed to exit the sides and bottom without reflection). A homogeneous Dirichlet boundary condition imposed on $\Gamma_\text{top} = \partial\Omega \cap \{\mathbf{x} \in \Omega | x_2 = 3\}$ has the effect of reflecting waves back. Our observations take place near the top boundary as would be done in an acoustic scattering inverse problem. This has the effect of complicating the observations due to the reflections. We use a $128 \times 128$ mesh and linear finite elements and linear basis functions for the parameter $m$, making $d_M = 16,641$ for this problem. We use quadratic basis functions for the state $u$. This problem has a complicated oscillatory response surface and demonstrates the benefit of AS basis for capturing complex mappings. 

Similar to the last problem, the dense DIPNet and PCANet architectures required more extensive optimization hyperparameter tuning to achieve desirable accuracy. We omit the comparison to these methods noting that both DIPResNet and PCAResNet uniformly outperformed DIPNet and PCANet in our numerical experiments. Instead for this problem we focus our comparison on a class of networks we were unable to train for the Poisson example due to the large input and output dimensions. In this case since $d_Q$ is 2 orders of magnitude smaller, we were able to reliably train fully-connected dense encoder-decoder networks. For this problem these networks have appropriate nonlinearity for capturing the oscillatory parameter-to-observable maps. We consider a ``full dense'' network that learns the full $16,641$ to $200$ mapping with $200$ dimensional hidden layers, and a ``truncated dense'' network that uses the same rank $r$ as the reduced basis ResNets for the hidden layer representations. As was stated prior, all networks have the same depths and use the same activation functions. 

\begin{figure}[H] 
\includegraphics[width = \textwidth]{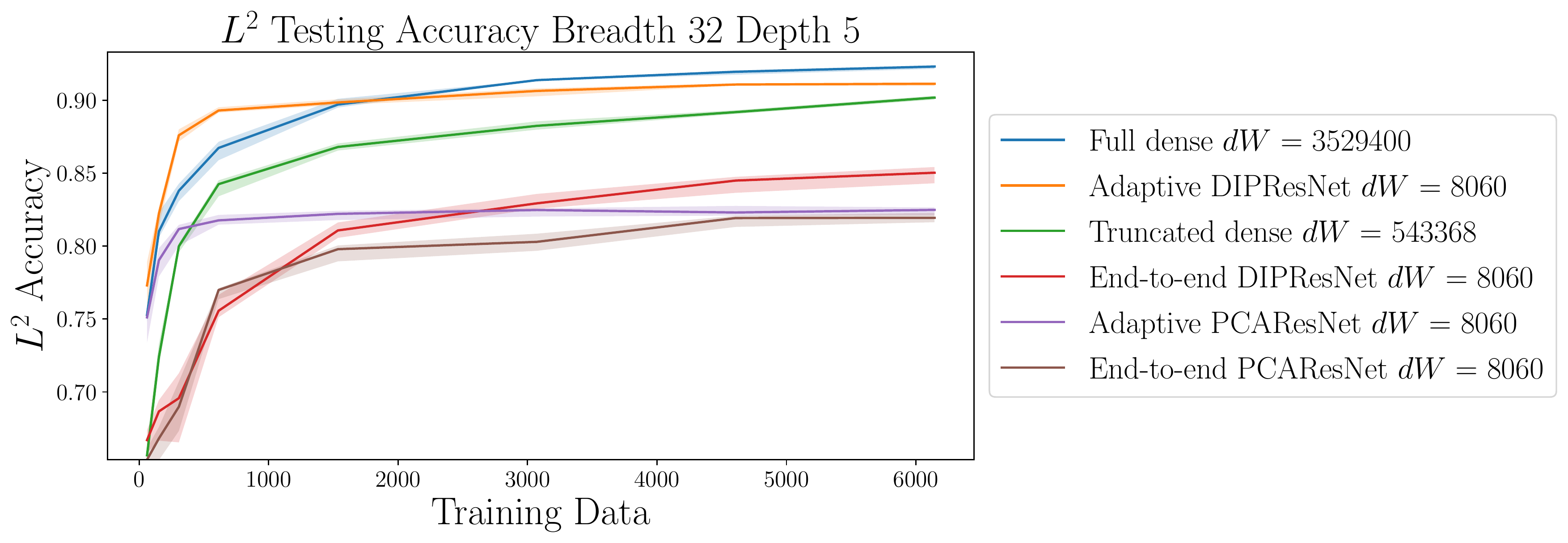}
\caption{Comparison of networks for depth = $5$, breadth = $32$. Since the input and truncation is significant for this problem the full dense network is able to outperform the reduced basis strategies, including the adaptive DIPResNet strategy, given enough data. The adaptively trained reduced basis ResNets outperform the end-to-end trained analogs, as in the last example. The DIPResNet significantly outperforms the PCAResNet in this case demonstrating the power of the AS reduced basis.  }
\label{helmholtz_depth_five_breadth32_comps}
\end{figure}

We begin by considering a case of depth $5$, starting with breadth $r=32$ and increasing it to enrich the basis function representation. Figure \ref{helmholtz_depth_five_breadth32_comps} shows that the adaptively trained DIPResNet outperformed all other models until the over-parametrized full dense model began modestly outperforming the breadth $32$ depth $5$ adaptively trained DIPResNet at around $1500$ training data; this is an expected result in machine learning: overparametrized networks \emph{should} have superior performance in the high data limit (note however that the full dense strategy is not scalable and will suffer with mesh refinement; see \cite{OLearyRoseberryVillaChenEtAl2022}). Additionally this Figure compares the performance of the DIPResNet (which used the active subspace basis) vs PCAResNet. This plot shows that while the two architectures initially have similar performance characteristics in the extremely low training data limit, the DIPResNet significantly outperforms the PCAResNet as more training data become available. Additionally, the adaptive PCAResNet outperforms the end-to-end trained DIPResNet until about $3,000$ training data. This plot demonstrates that \emph{both} the map informed reduced basis (AS) and the adaptive training strategy help improve accuracy when few training data are available.

\begin{figure}[H] 

\includegraphics[width = \textwidth]{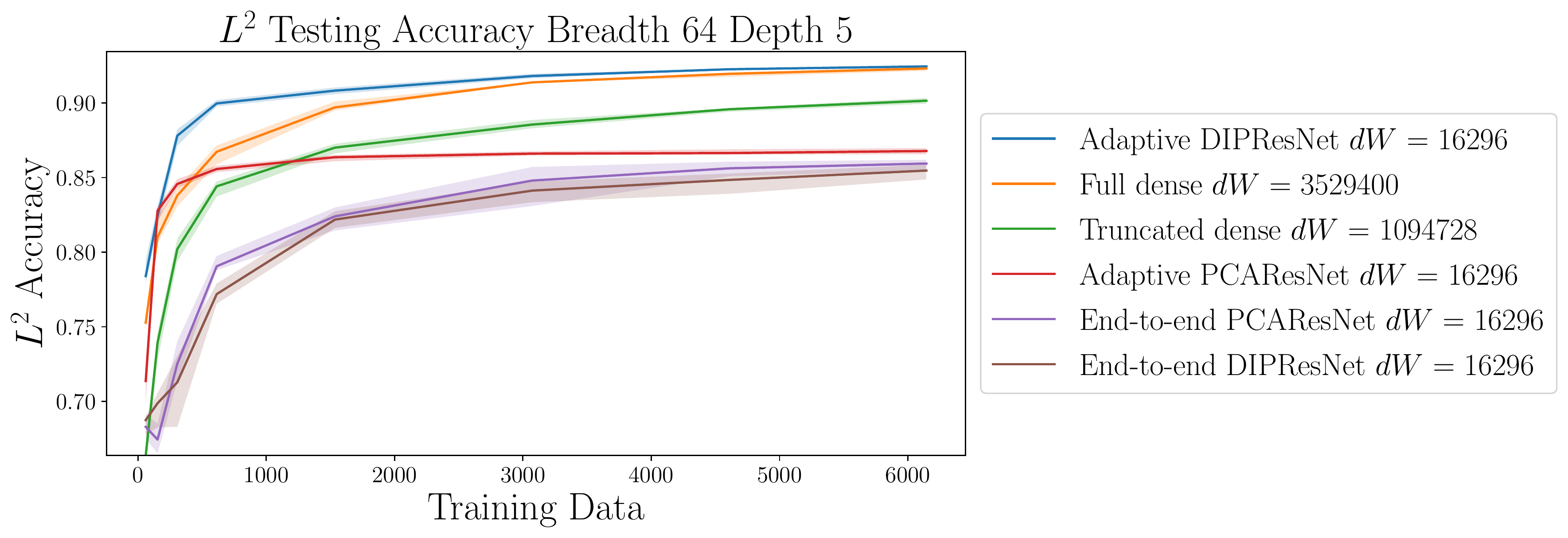}
\caption{Comparison of networks for depth = $5$, breadth = $64$. Doubling the reduced basis dimension for the adaptive DIPResNet allowed it to capture enough sensitive modes of the map to outperform the overparametrized full dense network. The truncated dense network was still unable to catch up to DIPResNet, but outperformed the PCAResNets. Similar results to Figure \ref{helmholtz_depth_five_breadth32_comps} hold where DIPResNet outperforms PCAResNet, and adaptive training outperforms end-to-end training. }
\label{helmholtz_depth_five_breadth64_comps}
\end{figure}

Since the output  of the Helmholtz problem is highly oscillatory we need larger reduced basis representations to faithfully capture $q$. Figure \ref{helmholtz_depth_five_breadth64_comps} demonstrates the effects of enriching both the input and output basis representations (as well as increasing the intermediate hidden neuron representations in the truncated dense network). As more basis functions are added to the inputs and outputs, the adaptively trained DIPResNet reliably outperforms the over-parametrized fully-connected dense network.

\begin{figure}[H] 
\includegraphics[width = \textwidth]{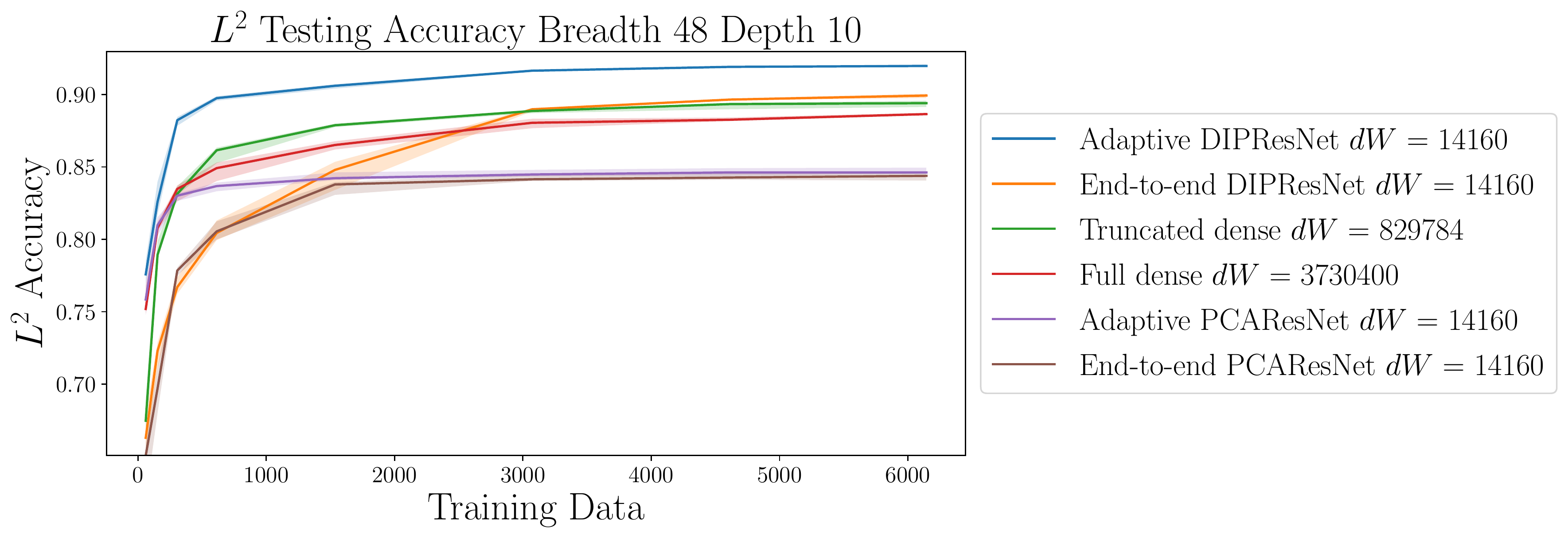}
\caption{Comparison of networks for depth = $10$, breadth = $48$. As depth is added to the network the full dense network becomes harder to train (the ``peaking phenomenon''), the truncated dense still performs reasonably well. The DIPResNet architectures were robust to the effects of adding depth, and gave good performance as more layers were added. Interestingly the end-to-end DIPResNet performed better than in the depth $5$ cases. The PCAResNet methods did not perform well, demonstrating the inability of the KLE basis to capture the map sensitivity adequately. }
\label{helmholtz_depth_ten_breadth48_comps}
\end{figure}

In the next set of numerical experiments, we study the effects of increasing the depth of the networks; in particular we are interested in robustness of the architectural strategies to added depth. In comparison to the depth $5$ networks, Figure \ref{helmholtz_depth_ten_breadth48_comps} demonstrates that the ResNet strategies are robust to adding more depth, while the full generic network tends to suffer. Interestingly the end-to-end trained ResNets are able to perform better as more depth is added, but the dense networks are not. 


As the numerical results demonstrate, the adaptive DIPResNet is well suited to this Helmholtz problem, along with the fully-connected dense encoder-decoder networks (i.e. full dense and truncated dense). We note however that, as the previous example demonstrated, when $d_Q,d_M$ get too large, fully-connected dense encoder-decoder strategies are no longer viable. The compressed adaptive ResNet can still produce good approximations in this case. For a similar Helmholtz problem in \cite{WuOLearyRoseberryChenEtAl22}, adaptively trained DIPResNets were capable of significantly accelerating the solution of Bayesian optimal experimental design sensor selection problems, producing accurate approximations of normalization constants for expected information gain computations. 



\subsection{Learning Inverse Aerodynamic Shape Optimization Maps}

In the previous two examples, the parametric surrogate can help accelerate the solution of complicated outer-loop problems by substituting the surrogate for a complex PDE solve in the inner loops. This strategy reduces the costs by cheapening the per-iteration cost significantly. Another, possibly bolder strategy is to attempt to \emph{learn the solution of the outer-loop problem} directly. In this example we define a parametric map by the solution of an aerodynamic shape optimization problem for the shape variables $q(m)$ as a function of inputs $m$ which represent design and flow constraints. The four inputs are a specified lift coefficient $C_L$, a specified bound on a moment coefficient $C_M$, a specified Reynolds number $Re$, and a lower bound for the volume, $V_\text{lb}$ in the interior of the 3D wing. The input parameters define a constrained optimization problem for the wing shape, which is stated as:

\begin{align*}
  \min_{q \text{ shape},\alpha \text{ AoA}} C_D(q,\alpha;M,Re) \nonumber\\
  \text{subject to}\\
  C_L(q,\alpha;,M,Re) = p_1\\
  C_M(q,\alpha;M,Re) \geq p_2 \\
  V(q,\alpha;M,Re) \geq V_\text{lb}
\end{align*}
\begin{figure}[H]
\begin{subfigure}{0.5\textwidth}
\includegraphics[width = \textwidth]{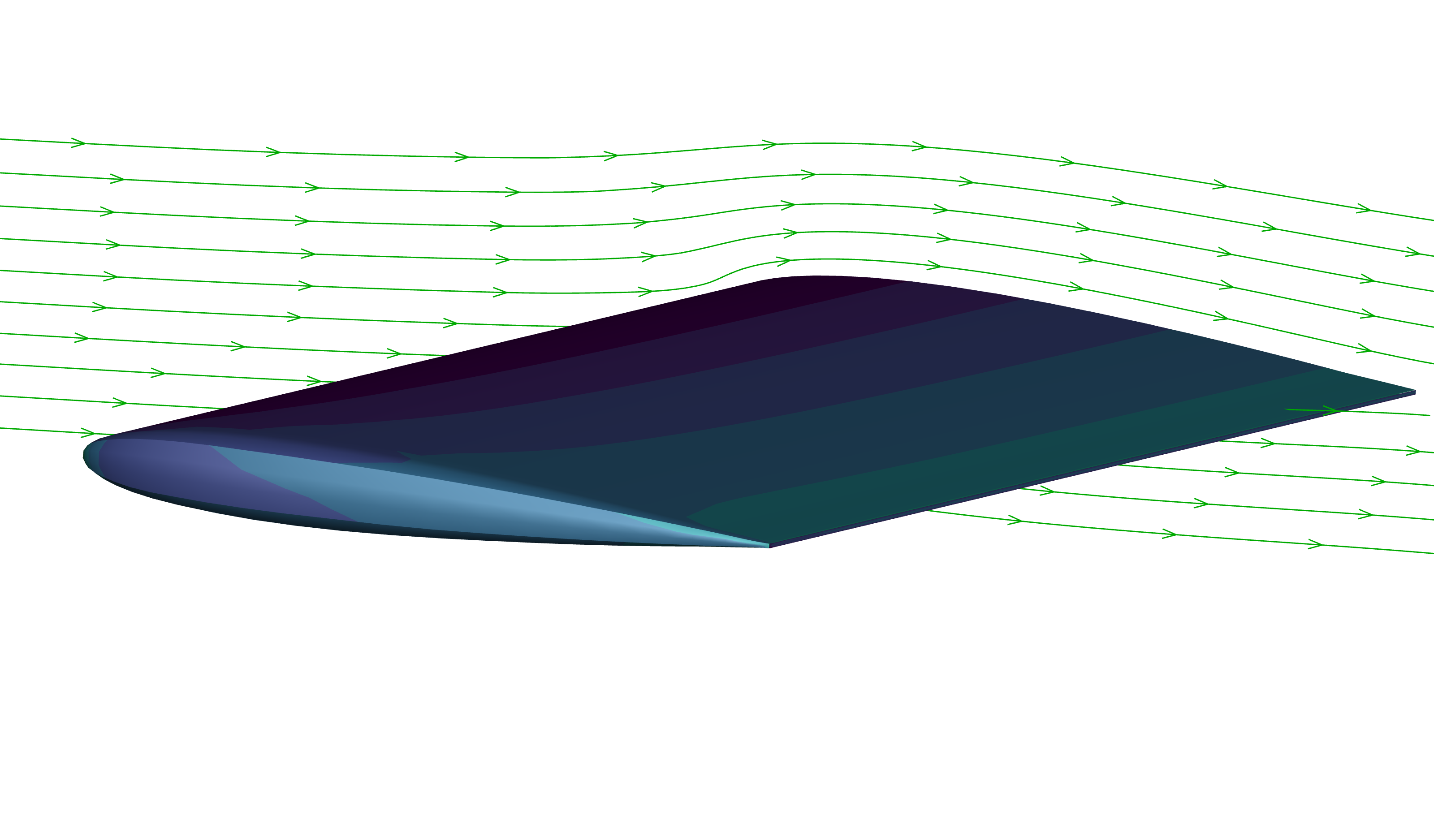}
\end{subfigure}%
\begin{subfigure}{0.5\textwidth}
\includegraphics[width = \textwidth]{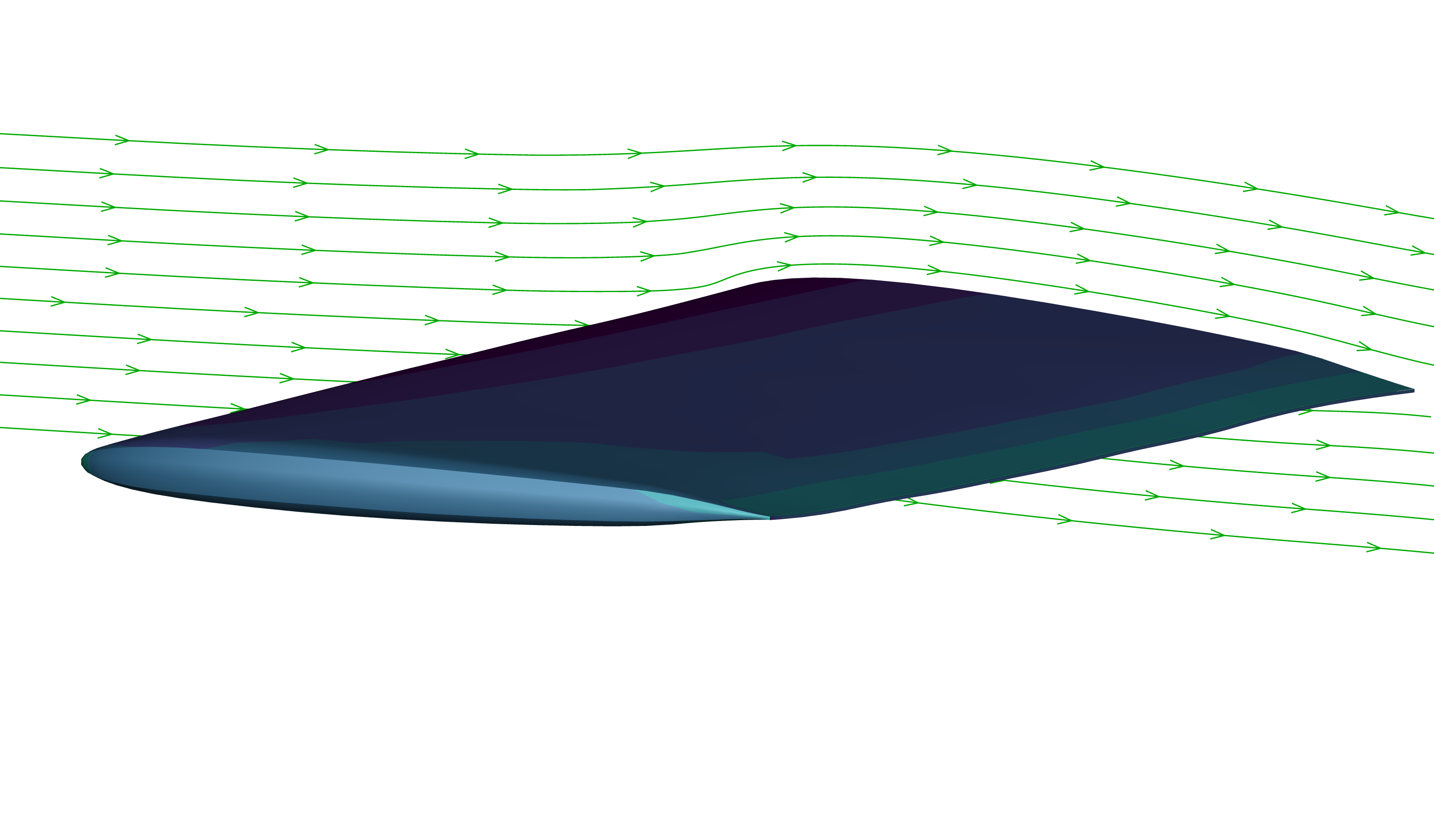}
\end{subfigure}

\caption{Left: baseline wing shape with streamlines, right: example optimal wing shape with streamlines}
\label{fig:aero_shape}
\end{figure}

The CFD optimization problem seeks to find the optimal shape $q$ and angle of attack that minimize the drag coefficient $C_D$ of the wing, subject to the constraints on $C_L,C_M,$ and the volume $V$. The scalar variables $C_D,C_L$ and $C_M$ depend on $Re,q$ and $\alpha$ through the solution of a Reynolds Averaged Navier-Stokes (RANS) CFD model. We sample the four input parameters $m=[p_1,p_2,Re,V_\text{lb}]$ from a uniform distribution that is based on a wing design problem defined by the Aerodynamic Design Optimization Discussion Group (ADODG) of the American Institute of Aeronautics and Astronautics (AIAA). The uniform intervals are $p_1 \in[0.2,0.4],p_2 \in [0.8V_\text{ref},1.0V_\text{ref}], and Re \in [10^5,10^7]$, where $V_\text{ref}$ is a reference lower bound defined in the ADODG problem formulation. The shape is represented by $200$ free form deformation (FFD) points that represent the smooth aerodynamic surface via spines, and additionally a $9$ dimensional twist parameter represents how different cross-sections of the wing rotate relative to a fixed initial frame of reference. In this work we only learn the $200$ dimension FFD points; the twist representation was easily captured by a small $4$-to-$9$ full-connected dense networks. For the problem we consider, the design constraints are low dimensional $d_M = 4$, and the design dimension is medium dimensional $d_Q = 200$; we demonstrate the usefulness of the adaptive reduced-basis ResNet strategy for problems of more modest dimensionality than the previously discussed PDE maps. 

A note on the significance of this example, and the corresponding accuracies. We observe the FFD points are a conservative quantity of interest; the accuracies in the optimal wing shape, and the optimal objective $C_D$ tend to be much better. In experiments a network that was $93.2\%$ accurate in the $L^2$ metric corresponded to an optimal wing shape that was $99.98\%$ accurate in the $C_D$ for data not seen during training. This came at a measured speedup of $~6\times 10^7$ over the 3D CFD shape optimization. For more information, we refer the reader to \ref{appendix:aerodynamic_shape_opt} for a more detailed discussion of the problem description. 

In this problem the input dimension is small so we do not use any input dimension reduction. Instead we reduce only the $200$ dimensional output, using POD. Thus the truncation errors in the analysis apply only to the output representation. For these numerical results we use a $r=10$ dimensional POD basis for the output representation. As a first step in the projected ResNet, we prolongate the $4$ dimensional inputs to the $10$ dimensional representation using a linear neural network layer ($10\times 4$ matrix with an affine shift), which is learned during training. The ResNet then learns the mapping between the prolongated input data representation and the $10$ dimensional output basis for POD. We refer to this network thus as PODResNet. We compare adaptively trained PODResNet to end-to-end trained PODResNet. We compare these strategies to fully-connected dense encoder-decoder networks which use the full $200$ POD dimension (``full dense'' as in the last example), and which use the same $10$ dimensional hidden layer representation as the PODResNet (``truncated dense'' as in the last problem). We study the effects of the size of the training data and the depth of the neural networks. We repeat runs over 20 different independent shuffles of a larger dataset of training data. We report median generalization accuracies $\pm 30\%$, as well as the maximum accuracies in some cases. We begin by studying a comparison of the different networks for the depth $5$ case. 

\begin{figure}[H] 

\includegraphics[width = \textwidth]{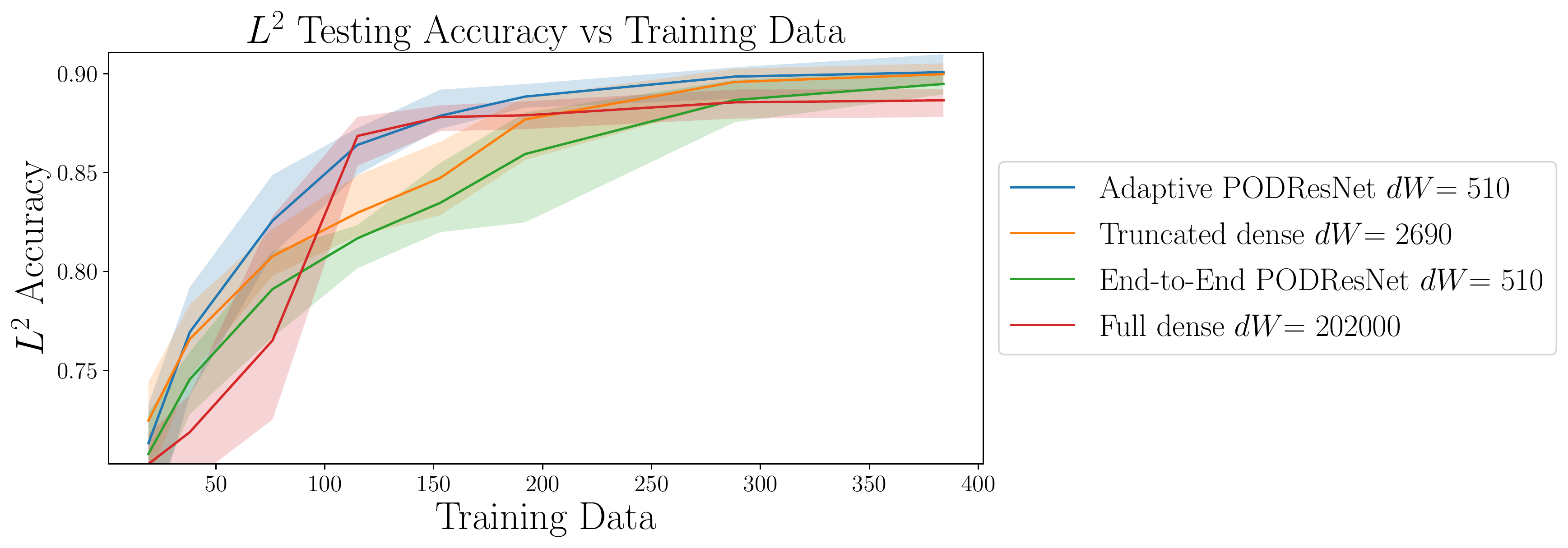}
\caption{Comparison of networks for depth = $5$.The adaptive ResNet strategy outperforms the dense strategies in the low data limit. However the full space dense network is able to perform about as well once around $100$ data become available. The truncated dense networks and end-to-end ResNet perform worse.}
\label{aero_depth_five_comps}
\end{figure}

\begin{figure}[H] 

\includegraphics[width = \textwidth]{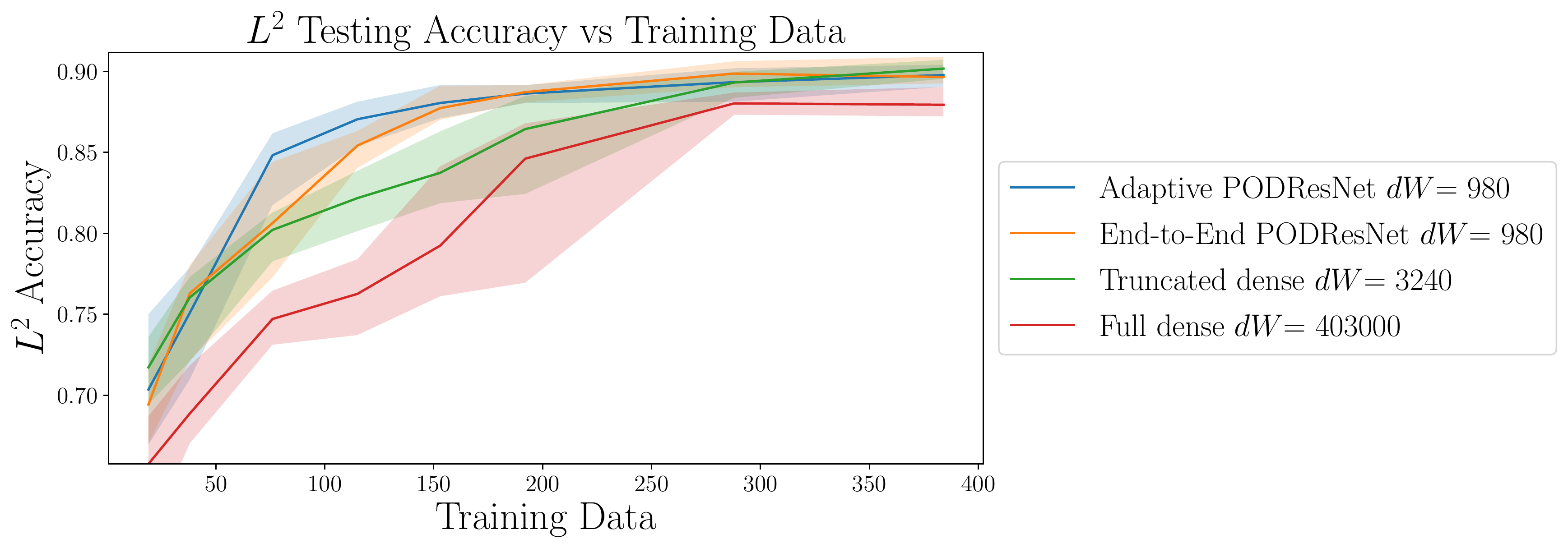}
\caption{Comparison of networks for depth = $10$. The adaptive PODResNet is still the best. Increasing the depth has the effects of deteriorating the performance of the full dense and truncated dense networks, as well as improving the performance of the end-to-end PODResNet.}
\label{aero_depth_ten_comps}
\end{figure}

\begin{figure}[H] 
\includegraphics[width = \textwidth]{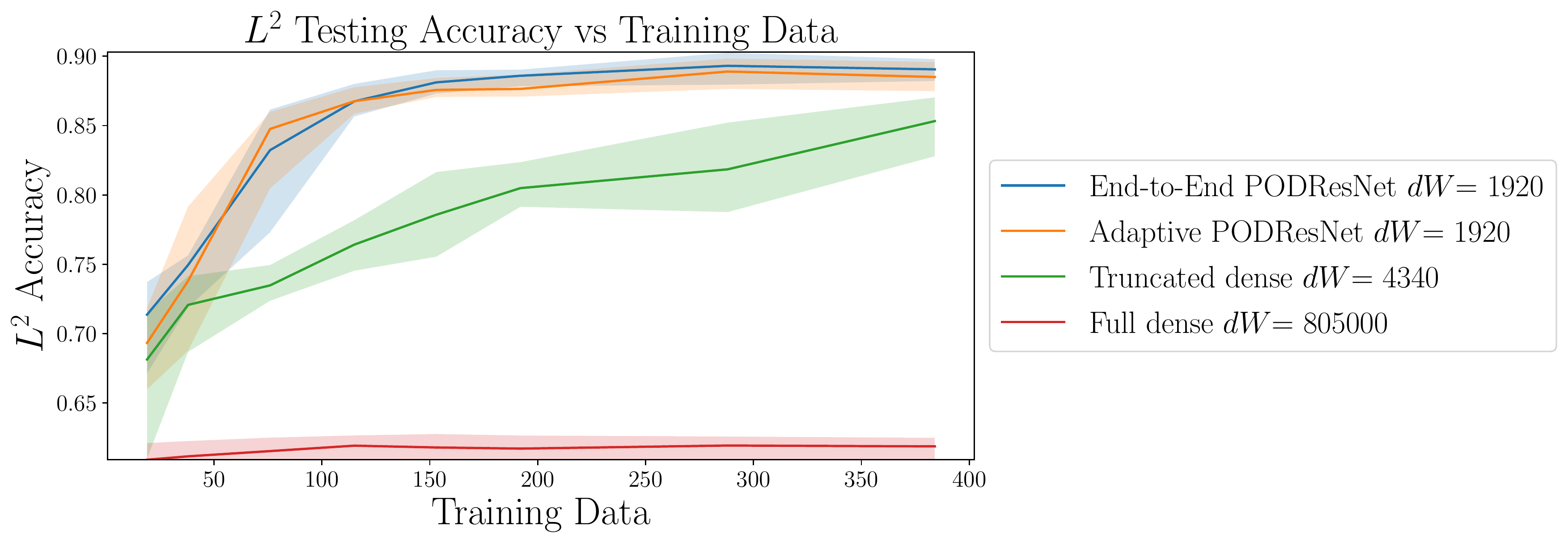}
\caption{Comparison of networks for depth = $20$. We observe the ``peaking phenomenon'' for FD; given limited data the $20$ layer fully-connected high-dimensional network is unable to yield generalization accuracy that similar architectures with less depth are able to achieve. }
\label{aero_depth_twenty_comps}
\end{figure}

Overall, the networks tend to perform more similarly to each other in these sets of example than in the previous PDE examples; we believe this is due to the relatively modest dimensionality of the problem. However the adaptively trained PODResNet is still able to outperform all of the other strategies. Figure \ref{aero_depth_five_comps} demonstrates that the adaptive PODResNet outperforms all of the other networks, and for modest depth, the truncated dense and end-to-end PODResNet performed worse than the full dense and adaptive PODResNet. Figure \ref{aero_depth_ten_comps} demonstrates that as we increase the depth the adaptive PODResNet still performs the best, but the end-to-end PODResNet performs better than in Figure \ref{aero_depth_five_comps}, as was also the case in previous examples. Figures \ref{aero_depth_ten_comps} and \ref{aero_depth_twenty_comps} together in comparison to Figure \ref{aero_depth_five_comps} demonstrates that the addition of depth significantly deteriorates the performance of the fully-connected dense encoder-decoder networks, while the adaptive ResNet architectures perform well, and are robust to the addition of depth.

These numerical results show that the adaptive ResNet strategy can outperform conventional black-box neural network strategies when limited data are available, which is critical because each data point amounts to solution of a full CFD-based shape optimization problem. Additionally the results show that the adaptive ResNets do not suffer from the degraded performance due to adding more depth that the dense feedforward networks do.

\section{Conclusion}

We have presented a constructive architectural framework for parametric deep learning in the form of adaptively trained reduced-basis ResNets. Appropriate linear compressibility of the high-dimensional map can be detected via reduced basis techniques in a constructive manner, equipped with rigorous approximation error bounds. The fundamental contribution of this work is to constructively build nonlinear approximations of the compressed map between the spaces spanned by the reduced bases, and to do so with good performance. By posing the reduced basis latent space learning problem as a sequential minimization problem, and equivalently as an architectural construction algorithm, we can achieve good approximation by deep networks, while simultaneously detect appropriate breadth and depth. 

The proposed methodology performed well in three different high-dimensional learning problems, and demonstrated the viability of parametric machine learning in very high dimensions, for problems that admit compressible structure. This is a critical need for outer-loop problems, that require many queries of $m\mapsto q$, a task that may be intractable in practice (such as the 3D CFD shape optimization presented here). When reasonable surrogate accuracies can be achieved for limited training data, the capabilities for solving outer-loop problems can be extended significantly. 

Conventional wisdom in machine learning suggests that the best strategies for approximating input-output mappings is to build overparametrized neural networks, which may have much larger weight dimensions than the available training data cardinality. The associated large configuration spaces give the optimizer more freedom to avoid spurious stationary points, and better fit the data. The typical setting for machine learning, however, is ``big-data,'' where large training datasets are available for the empirical risk minimization problem (or the model is already initialized well, e.g., transfer learning). In the setting that we are concerned with, the opposite is seen to hold, that when one can afford just a few samples of a very high dimensional map, it can be a liability to overparametrize. 

A last advantage worth noting is the computational economy of our proposed approach. Improving the computational economy of neural networks, (e.g., ``pruning'' \cite{BlalockOrtizFrankleEtAl2020state}) is an issue of recent concern. In order to scale machine learning models to energy and memory constrained environments, parsimonious network surrogates are needed. This approaches attempts to build an already ``pruned'' model from knowledge of the structure of the input-output map.

\section*{Acknowledgment}
We are grateful to the anonymous reviewers, whose comments helped improve this work significantly.

\appendix

\section{Analysis}

\subsection{Monte Carlo Bound for Sample Approximation of $\mathbb{E}_\nu[AA^T]$} \label{AAT_mc_bound}

For generality suppose $A_iA_i^T \in \mathbb{R}^{n \times n}$ and $ x \in \mathbb{R}^{n}$ is arbitrary, then via Jensen's inequality and other bounds we have

\begin{align}
\left(\mathbb{E}_{m_i \sim \nu}\left[\left\|\frac{1}{N}\sum_{i=1}^N A_iA_i^Tx - \mathbb{E}_\nu[AA^T]x \right\|_2\right]\right)^2 \nonumber \\
 \leq \mathbb{E}_{m_i \sim \nu}\left[\left\|\frac{1}{N}\sum_{i=1}^N A_iA_i^Tx - \mathbb{E}_\nu[AA^T]x \right\|^2_2\right] \nonumber \\
= \text{tr} \left(\text{Cov}\left(\frac{1}{N}\sum_{i=1}^N A_iA_i^Tx - \mathbb{E}_\nu[AA^T]x\right)\right) \nonumber \\
\leq \frac{1}{N^2}\times N \text{tr} \left(\text{Cov}\left( A_iA_i^Tx - \mathbb{E}_\nu[AA^T]x\right)\right) \nonumber \\
=\frac{1}{N}\left\|\mathbb{E}_{m_i \sim \nu}\left[\left(A_iA_i^T - \mathbb{E}_\nu[AA^T] \right)x\right]\right\|_2^2  \leq \frac{1}{N} \sigma_A^2 \|x\|^2_2.
\end{align}

By taking a supremum over all $x \in \mathbb{R}^{n}$ such that $\|x\|_2 = 1$ we get the following result for the matrix norm:
\begin{equation}
    \left(\mathbb{E}_{m_i \sim \nu}\left[\left\|\frac{1}{N}\sum_{i=1}^N A_iA_i^T - \mathbb{E}_\nu[AA^T] \right\|_2\right]\right) \leq \frac{\sigma_A}{\sqrt{N}}.
\end{equation}

\subsection{Proof of Proposition \ref{prop_basis_error}} \label{proof_prop_basis_error}

We begin by making use of the triangle inequality:
\begin{align} \label{three_term_split_mcmc_rb_bound}
    \|q - \widetilde{\Phi}_r \circ q_{r}\circ \widetilde{V}_{r}^T\|_{L^2_\nu} \leq \|q - \Phi_r \circ q_{r}\circ V_{r}^T\|_{L^2_\nu} + \nonumber \\
     \|\Phi_r \circ q_{r}\circ V_{r}^T - \widetilde{\Phi}_r \circ q_{r}\circ V_{r}^T\|_{L^2_\nu} + \|\widetilde{\Phi}_r \circ q_{r}\circ V_{r}^T - \widetilde{\Phi}_r \circ q_{r}\circ \widetilde{V}_{r}^T\|_{L^2_\nu}.
\end{align}
The first of the three terms is bounded by $\epsilon_\text{ridge}$ by assumption. The ridge function itself can then be bounded by the reverse-triangle inequality as
\begin{equation}
    \|\Phi_r \circ q_r \circ V_r^T\|_{L^2_\nu} \leq \|q\|_{L^2_\nu} + \epsilon_\text{ridge}.
\end{equation}

Since $\Phi_r$ is orthonormal we can bound in the reduced space:
\begin{equation}
\|q_r \circ V_r^T\|^2_{\L^2(\mathbb{R}^{d_M},\nu;\mathbb{R}^{r})}= \|\Phi_r \circ q_r \circ V_r^T\|_{L^2_\nu} \leq \|q\|_{L^2_\nu} + \epsilon_\text{ridge}.
\end{equation}
Then the second term in \eqref{three_term_split_mcmc_rb_bound} can be bounded using the operator norm:
\begin{align}
    &\|\Phi_r \circ q_{r}\circ V_{r}^T - \widetilde{\Phi}_r \circ q_{r}\circ V_{r}^T\|_{L^2_\nu} = \|\Phi_r - \widetilde{\Phi}_r\|_{\ell^2(\mathbb{R}^{d_Q \times r})}\|q_r \circ V_r^T\|_{L^2(\mathbb{R}^{d_M},\nu;\mathbb{R}^{r})} \nonumber\\
    &\leq \epsilon_\text{output}\|q_r \circ V_r^T\|_{L^2(\mathbb{R}^{d_M},\nu;\mathbb{R}^{r})}\leq \epsilon_\text{output} (\|q\|_{L^2_\nu} + \epsilon_\text{ridge}).
\end{align}
Finally, the third term in \eqref{three_term_split_mcmc_rb_bound} can be bounded as follows:
\begin{align}
    &\|\widetilde{\Phi}_r \circ q_{r}\circ V_{r} - \widetilde{\Phi}_r \circ q_{r}\circ \widehat{V}_{r}\|_{L^2_\nu}\nonumber \\
    &= \|\widetilde{\Phi}_r\|_{\ell^2(\mathbb{R}^{d_Q \times r})}\sqrt{\int \| q_r(V^T_{r_M}m) - q_r(\widehat{V}^T_{r_M}m)\|_{\ell^2(\mathbb{R}^r)}^2 d\nu(m)} \nonumber \\
    &\leq L_{q_{r_M}} \sqrt{\int \| V^T_{r_M}m - \widehat{V}^T_{r_M}m\|_2^2 d\nu(m)} \nonumber \\
    &= L_{q_{r_M}} \| V^T_{r_M} - \widehat{V}^T_{r_M}\|_{\ell^2(\mathbb{R}^{r \times d_M})}\sqrt{\int \| m\|_2^2 d\nu(m)} \nonumber  \\
    & \leq L_{q_{r_M}} \epsilon_\text{input} \left(\sqrt{\int \| m - m_0\|_2^2 d\nu(m)} +\sqrt{\int \| m_0\|_2^2 d\nu(m)}\right) \nonumber \\
    & = L_{q_{r_M}} \epsilon_\text{input} \left[\sqrt{\text{tr}(\mathcal{C})} + \|m_0\|_2 \right].
\end{align}
The total bound follows.

\subsection{Proof of Theorem \ref{projected_resnet_representation_theorem}} \label{representation_error_appendix}

Let $\epsilon>0$ be arbitrary, and let $\chi_K$ denote the characteristic function for the compact set $K \subset\subset \mathbb{R}^{d_M}$. By the reverse triangle inequality we have that $\Phi_r \circ q_r \circ V_r^T \in L^2(\mathbb{R}^{d_M},\nu;\mathbb{R}^{d_Q})$:
\begin{equation}
    \left|\|\Phi_r \circ q_r \circ V_r^T\|_{L^2_\nu} - \|q \|_{L^2_\nu} \right| \leq \|\Phi_r \circ q_r \circ V_r^T - q \|_{L^2_\nu} \leq \zeta,
\end{equation}
so 
\begin{equation}
    0 \leq \|\Phi_r \circ q_r \circ V_r^T\|_{L^2_\nu}  \leq \|q \|_{L^2_\nu} + \zeta < \infty.
\end{equation}

Since $\|\Phi\|_{\ell^2(\mathbb{R}^{d_Q\times r})} = 1$, we have that $q_r \circ V_r^T \in L^2(\mathbb{R}^{d_M},\nu;\mathbb{R}^r)$. In $L^2(\mathbb{R}^{d_M},\nu;\mathbb{R}^r)$ we can arbitrarily well approximate $q_r \circ V_r^T$ by a continuous function $q_r^\text{cont} \in C^0(\mathbb{R}^{d_M},\nu; \mathbb{R}^r)$ when restricted to $K$. Let $q_r^\text{cont}$ be such that $\mathbb{E}_{\nu}[ \| q_r - q_r^\text{cont}\|^2_{\ell^2(\mathbb{R}^{r})} \chi_{K}] < \frac{\epsilon^2}{9}$.

Proposition 4.11 in \cite{LiLinShen2019} states that any continuous function $q^\text{cont}_r \in C(\mathbb{R}^{d_M};\mathbb{R}^r)$ can be approximated arbitrarily well in $L^p(\mathbb{R}^{d_M};\mathbb{R}^{r})$, for $p\geq 1$, by a finite time control flow representation as long as the set of right hand sides $\mathcal{F}$ for the control flow is closed under affine operations, and the closure of this set under the topology of compact convergence contains a well function. The restricted affine invariance requires that if $f \in \mathcal{F}$, then $g(x) = Df(Ax +b)$ is also in $\mathcal{F}$, for $D,A$ diagonal matrices in $\mathbb{R}^{r\times r}$ with diagonal entries $d_i = \pm 1$ and $a_i\leq 1$, and $b \in \mathbb{R}^{r}$ arbitrary. We note that the family of right hand sides associated with the continuous analog of ResNet satisfies this property. The well function property requires that the activation function used in the ResNet can be used to build a function that is arbitrarily close to zero when restricted to an open bounded set (e.g., ReLU, sigmoid, tanh, etc.). For a lengthier discussion of these requirements see Section 2 in \cite{LiLinShen2019}. Proposition 4.11 along with these properties establishes that there exists a finite time $T<\infty$ control flow mapping of the form:
\begin{subequations}\label{resnet_control_flow_mapping}
\begin{align} 
	\frac{dz}{dt}(t) &= w_1(t)\sigma(w_0(t) + b(t)) \label{resnet_control_ode_appendix}\\
	z(0) &= V_r^T m\\
	\xi_T(m) &= z(T;m)
\end{align}
\end{subequations}
such that $\int_{K} \|\xi_T - q^\text{cont}_r\|^2_{\ell^2(\mathbb{R}^{r})}dm$ is arbitrarily small. Additionally we may require that the activation function has one derivative (e.g., tanh, softplus, sigmoid). In the interest of extending this result from $L^2(\mathbb{R}^{d_M};\mathbb{R}^r)$ to $L^2(\mathbb{R}^{d_M},\nu;\mathbb{R}^r)$ we employ the fact that the probability density function $\pi(m)$ for the measure $\nu$ is assumed to be essentially bounded. In other words $d\nu(m) = \pi(m)dm$ and there exists $C_\pi$ such that the measure of the set $\{m \in \mathbb{R}^{d_M}| \pi(m) > C_\pi\}$ is zero. For convenience we require that $\int_{K} \|\xi_T - q^\text{cont}_r\|^2_{\ell^2(\mathbb{R}^{r})}dm < \frac{\epsilon^2}{9C_\pi}$. The essential boundedness of $\pi(m)$ allows us to extend this result from $L^2(\mathbb{R}^{d_M};\mathbb{R}^{d_Q})$ to $L^2(\mathbb{R}^{d_M},\nu;\mathbb{R}^{d_Q})$, and at the same time obtain a bound by $\frac{\epsilon^2}{9}$:
\begin{align}
	&\int_K \|\xi_T - q^\text{cont}_r\|^2_{\ell^2(\mathbb{R}^{r})}\pi(m) dm \leq C_\pi\int_K \|\xi_T - q^\text{cont}_r\|^2_{\ell^2(\mathbb{R}^{r})}dm \leq \frac{\epsilon^2}{9}. 
\end{align}

The system \eqref{resnet_control_flow_mapping} can be approximated to arbitrary precision via an explicit Euler discretization with time step $\Delta t$, which yields the ResNet:
\begin{equation}
 	z_{k+1} = z_k + \Delta t w_{1k} \sigma(w_{0k}z_k + b_k)
\end{equation} 
Assuming the right hand side  of \eqref{resnet_control_ode_appendix} is Lipschitz with bound $L_\text{ResNet}$, and the true solution $z(t;m)$ to \eqref{resnet_control_flow_mapping} is itself twice differentiable for all $t \in (0,T)$, and for all $m \in K$ we have that $\max_{t\in(0,T]} \left\|\frac{\partial^2 z}{\partial t^2}\right\|_{\ell^2(\mathbb{R}^{r})} \leq M < \infty$, then the global truncation error for the explicit Euler approximation can be bounded by
\begin{equation}
	\int_K\|\xi^{E.E.}_T(m) - \xi_T(m)\|^2_{\ell^2(\mathbb{R}^r)} \pi(m)dm \leq \left(\frac{e^{TL_\text{ResNet}}-1}{2L_\text{ResNet}} M \Delta t\right)^2 C_\pi |K| \leq \frac{\epsilon^2}{9};
\end{equation}
see \cite{Quarteroni2010SaccoSaleri}. The requirement for $\Delta t$ become:
\begin{equation}
	\Delta t \leq \frac{2}{3}\frac{\epsilon L_\text{ResNet}}{M(e^{TL_\text{ResNet}}-1)\sqrt{C_\pi |K|}}.
\end{equation}
For homogenous time steps we have $\Delta t = \frac{T}{\text{depth}}$, which gives us the following bound:
\begin{equation}
	\text{depth} \geq \frac{3}{2} \frac{MT(e^{TL_\text{ResNet}}-1)\sqrt{C_\pi|K|}}{\epsilon L_\text{ResNet}}.
\end{equation}
Combining all of these results via the triangle inequality we have:
\begin{align}
	&\sqrt{\mathbb{E}_{\nu}\left[\|\xi^{E.E.}_T - q_r\|^2_{\ell^2(\mathbb{R}^r)}\chi_K\right]} \leq \nonumber \\
	&\sqrt{\mathbb{E}_{\nu}\left[\|\xi^{E.E.}_T - \xi_T \|^2_{\ell^2(\mathbb{R}^r)}\chi_K\right]} + \cdots \nonumber \\ 
    &\sqrt{\mathbb{E}_{\nu}\left[\|\xi_T - q_r^\text{cont} \|^2_{\ell^2(\mathbb{R}^r)}\chi_K\right]}+\sqrt{\mathbb{E}_{\nu}\left[\|q_r^\text{cont}- q_r \|^2_{\ell^2(\mathbb{R}^r)}\right]}  \leq \nonumber\\
	&\frac{\epsilon}{3} + \frac{\epsilon}{3} + \frac{\epsilon}{3}
\end{align}

The final result comes from setting $\epsilon = \zeta_r$ (since it was arbitrary), and then noting that $\xi_T^{E.E.}(m,w_T) = f_{w,r}(m,w_T)$,
\begin{align}
  	&\sqrt{\int_K\|q(m) - \Phi_r f_{w,r}(V_r^Tm,w_T)\|^2_{\ell^2(\mathbb{R}^{d_Q})} d\nu(m)} \leq \nonumber \\
  	&\sqrt{\int_K\|q(m) - \Phi_r q_r(V_r^Tm,w_T)\|^2_{\ell^2(\mathbb{R}^{d_Q})}d\nu(m)} + \cdots \nonumber \\
    &\sqrt{\int_K\| \Phi_r (q_r(V_r^Tm) - f_{w,r}(V_r^Tm,w_T) )\|^2_{\ell^2(\mathbb{R}^{d_Q})} d\nu(m)}
\end{align}
and 
\begin{align}
	\| \Phi_r (q_r(V_r^Tm) - f_{w,r}(V_r^Tm,w) )\|^2_{\ell^2(\mathbb{R}^{d_Q})} \leq \nonumber \\
	\|\Phi_r\|^2_{\ell^2(\mathbb{R}^{d_Q \times r})}\|q_r(V_r^Tm) - f_{w,r}(V_r^Tm,w) \|^2_{\ell^2(\mathbb{R}^{r})} \leq \nonumber\\
	\|q_r(V_r^Tm) - f_{w,r}(V_r^Tm,w) \|^2_{\ell^2(\mathbb{R}^{r})}
\end{align}
since $\Phi_r$ is orthonormal.

Note additionally that the $e^T$ complexity in the depth comes only from the global truncation error for explicit Euler, which is a conservative bound. In \cite{LiLinShen2019} the control flows they construct are actually piecewise constant in time, which is sufficient since simple functions are dense in continuous functions. In that case explicit Euler is an exact time integrator, and then the depth complexity can be reduced to $O\left(\frac{T|K|}{\epsilon}\right)$. The bound we give above, however, is more general since it represents discretizations of continuous time control flows.






\section{Learning Aerodynamic Shape Optimization Maps} \label{appendix:aerodynamic_shape_opt}

In this section, we provide details on the aerodynamic wing design problem defined by the Aerodynamic Design Optimization Discussion Group (ADODG) of the American Institute of Aeronautics and Astronautics (AIAA).
The AIAA ADODG formulates a series of benchmark cases that provide a foundation for rational assessment of the numerous aerodynamic design optimization approaches to problems of interest.
In particular, we generalize the ADODG Case 3, i.e., drag minimization of a rectangular wing in inviscid subsonic lifting flow, in the following ways:
we generalize the inviscid flow to viscous flow, which is a more challenging test case and requires solution of the Reynolds-averaged Navier–Stokes (RANS) equations; and we incorporate free-form deformation (FFD) control points as design variables as well as a twist angle distribution.
To parametrize the aerodynamic shape optimization problem, we expand each of the flight conditions and design requirements from a single value to a uniform parameter distribution $\nu(m)$ delineated in Table \ref{tab:opt}; this then defines our task for parametric regression of the mapping from parameter inputs $m$ to the optimal design variables $q(m)$.

\subsection{Problem Description}

\begin{figure}[!htb]
  \centering
  \includegraphics[width=0.8\textwidth,trim = {0, 2cm, 0, 2cm}, clip]{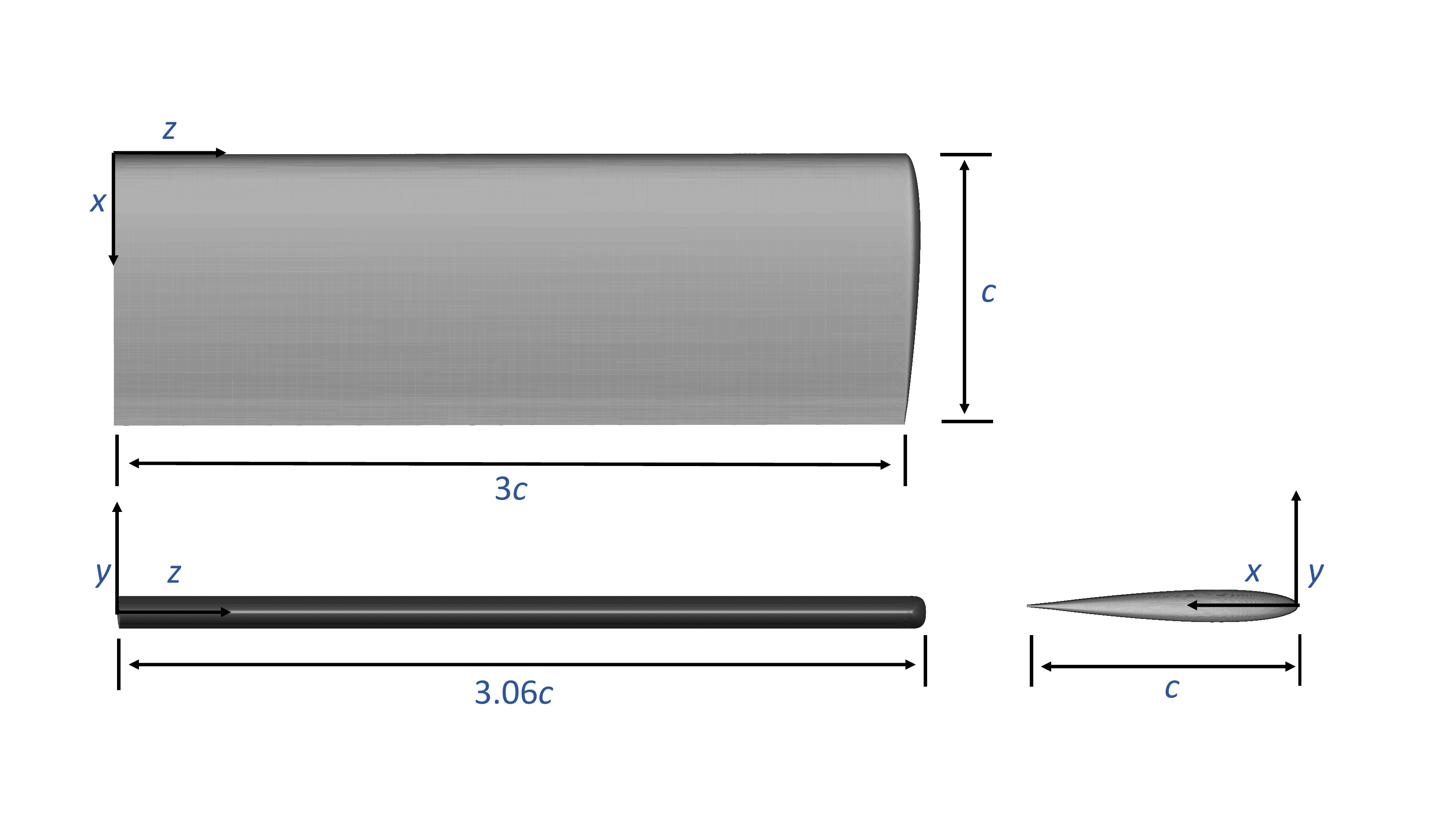}
  \caption{The rectangular wing has a chord of $c=1$, half wingspan of $3c$, a wing-tip cap of $0.06c$, and no twist angle, dihedral angle or sweep angle.}
  \label{fig:wingGeo}
\end{figure}

We set up the aerodynamic design problem by following the ADODG Case 3, drag minimization of a rectangular wing~\ref{fig:wingGeo} with the NACA 0012 airfoil as the initial guess for the wing section in inviscid subsonic flow.
For the purpose of testing the proposed algorithm in practical and challenging applications, we consider viscous flow by solving the RANS equations.
We consider 200 FFD control points to parameterize the wing geometry, with each of 10 wing sections having 10 FFD control points on the upper surface and 10 on the lower surface.
In addition, we use 10 twist variables, which are evenly distributed along the span-wise direction within the range of $[0.0, 3.0]$. The angle of attack is used as a dummy variable to satisfy the target lift coefficient constraint.
In this section we focus on learning the map for the FFD control points only, for which the POD basis admits low rank structure. Due to the infinite-dimensional nature of the wing shape, the twist and angle of attack are very low dimensional and do not lend themselves well to dimension reduction; a shallow dense neural network is sufficient for that task.
Since the flight condition is at low speed, varying Mach number will not significantly affect the aerodynamic performance.
Therefore, the Mach number is not considered as an input parameter.
We set the target lift coefficient within the range of $[0.2, 0.4]$, vary the lower bound of moment coefficient by $\pm$ 20\% to define a range of $[-0.11, -0.074]$, set the lower bound of internal volume as $0.8 \sim 1.0$ with respect to the baseline, and permit the Reynolds number to vary within the range of [$1\times10^5$, $1\times10^7$]. 
We summarize the wing design problem below in Table~\ref{tab:opt}.

\begin{figure}[h!]
    \begin{subfigure}{0.44\textwidth} {
    \includegraphics[width=\textwidth]{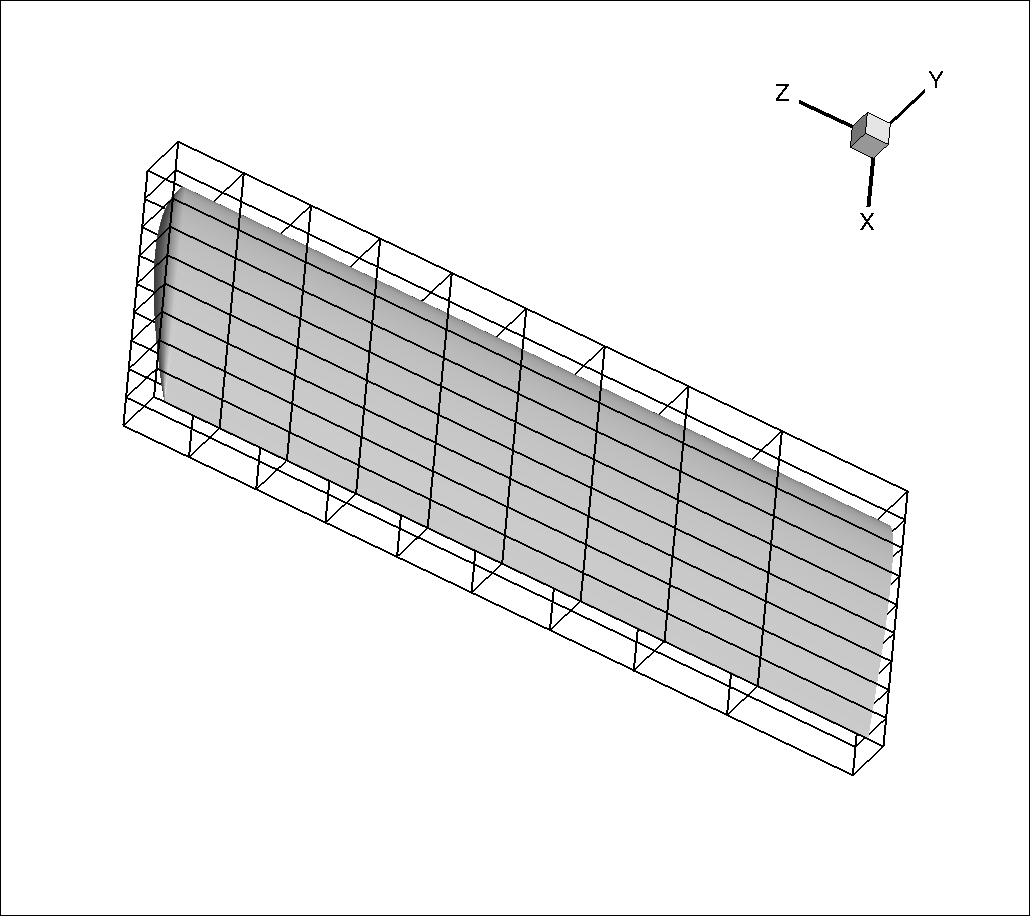}}
    \caption{FFD box used for optimization.}
    \end{subfigure}
    \begin{subfigure}{0.54\textwidth} {
    \includegraphics[trim= 100 00 120 00, clip, width=\textwidth]{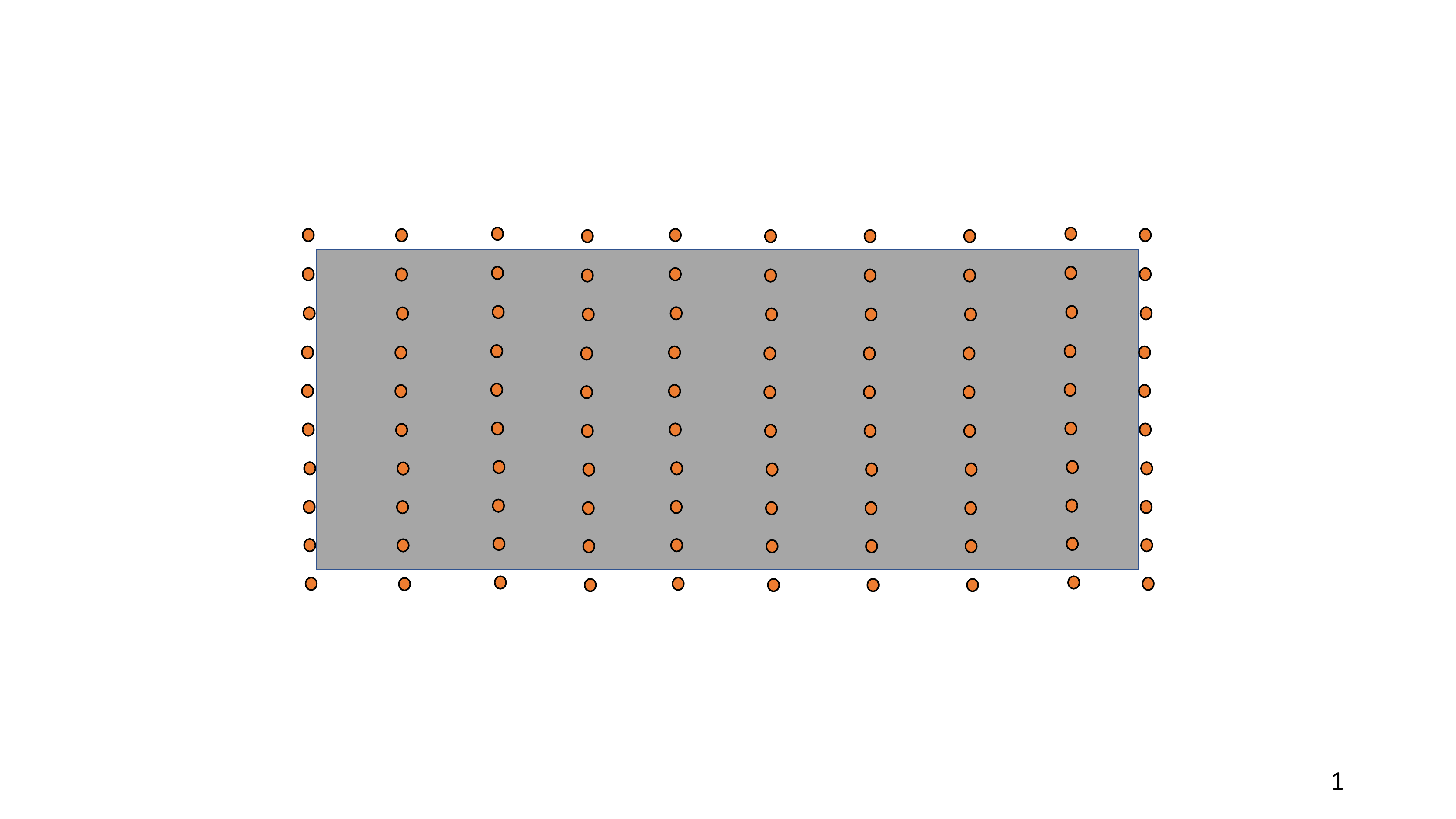}}
    \caption{FFD control points.}
    \end{subfigure}
\caption{A total of 200 (10 sections $\times$ 20 FFD control points / section) geometric variables provide sufficient design flexibility.}
\label{fig:ffd}
\end{figure}

\begin{figure}[h!]
\begin{center}
    \begin{subfigure}{0.6\textwidth} {
    \includegraphics[width=\textwidth]{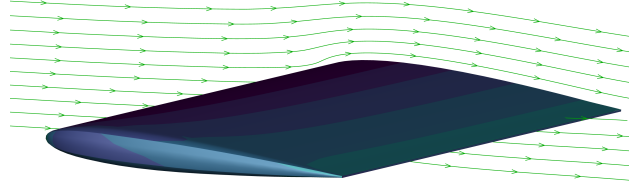}}
    \caption{Baseline airfoil}
    \end{subfigure}\\
    \begin{subfigure}{0.6\textwidth} {
    \includegraphics[trim= 100 00 120 00, clip, width=\textwidth]{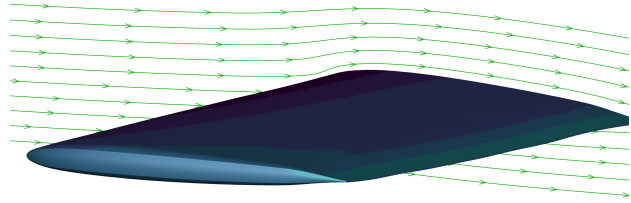}}
    \caption{Optimal airfoil for given constraints}
    \end{subfigure}
\end{center}
\caption{Comparison of the flow field for a baseline wing (left) and optimal wing (right) for a given set of constraints}
\label{fig:baseline_optimal}
\end{figure}

\begin{table*}[h!]
  \small
  \centering
  \caption{\small 3D wing optimization problem formulation that was used to help define the range of design parameters used in the learning problem. Note that we only learn the 200 dimensional FFD output.
  }
  \label{tab:opt}
  \begin{tabular}{lllr}
  \hline
          &  Function or variable   &  Description   &  Quantity \\ \hline
  minimize   &  ${C}_D$ &  Drag coefficient  &   \\
  w.r.t. \rule{0pt}{4ex}  & $\bold{x}$ & FFD control points  &  200  \\
                & $\boldsymbol{\lambda}$ & Twist angle & 9 \\
                    & $\alpha$ &  Angle of attack  &  1  \\
                  &  & {\bf Total design variables}  &  {\bf 210}  \\
  subject to \rule{0pt}{4ex}
               & ${C}_L^*$ = 0.2625 &  Lift-coefficient constraint &  1  \\
               & ${C}_M$ \textgreater \ -0.092 &  Moment-coefficient constraint &  1  \\
               & $0.8 V_0 \le V \le 1.5 V_0$ & Internal volume constraint &  1  \\
               & $0.8 \mathbf{t}_\text{base} \le \mathbf{t} \le 1.5 \mathbf{t}_\text{base}$ & Thickness constraints & 100\\
               &    & {\bf Total constraints} &  {\bf 103}  \\
  Conditions \rule{0pt}{4ex}  & $M = 0.5$ & Mach number &    \\ %
                    & $Re$ = $1 \times 10^6$ &  Reynolds number  &   \\ 
  \hline
  \end{tabular}
\end{table*}

\subsection{MACH-Aero Design Framework} \label{mach_aero_appendix}

MACH design framework targets Multidisciplinary design optimization of Aircraft Configurations with High fidelity while MACH-Aero~\footnote{\url{https://github.com/mdolab/MACH-Aero}} implements MACH on aerodynamic design optimization problems.
MACH-Aero sets up an optimization problem using the \texttt{python} interface pyOptSparse~\cite{Wu2020} and starts with an initial design $q_0$ under specified design requirements $m$, and uses a gradient-based quasi-Newton method to find the optimum airfoil design (control points and angle of attack). 
The steps are as follows. 
(1) A baseline design volume mesh is generated using pyHyp~\cite{Secco2021}, which will be deformed for any given value of design variables.  
(2) The SNOPT (Sparse Nonlinear OPTimizer)~\cite{Gill2002a} updates design variables and sends the new design to the geometry parameterization.
(3) The geometry parameterization module (pyGeo)~\cite{Kenway:2010:C} performs the geometry deformation, and computes the values of geometric constraints and corresponding gradients. 
(4) The volume mesh deformation module generates the deformed mesh based on the deformed geometry. 
(5) The CFD module (ADflow~\cite{Mader2020a} or DAFoam~\cite{He2020b}) computes high fidelity forward and adjoint flow fields on the deformed mesh, and sends the objective function and constraint values and computed gradient information back to the optimizer. The process is iterated until an optimal design $q(m)$ is found that satisfies the optimality conditions.

\section{Two-Step Optimization for Neural Network Training} \label{appendix:two_step_opt}

In this section we provide information on the two-step optimization procedures that we use in Section \ref{section_numerical_experiments}. We first use an Adam optimizer with default settings for a small number of epochs for each adaptive layer training (total epochs sums to $50$), and then perform one final end-to-end training with LRSFN \cite{OLearyRoseberryAlgerGhattas2020} for $50$ ``epoch equivalent'' neural network sweeps (i.e., forward and backward pass). In this case we allow all intermediate ResNet layers (and the output layer) to be trained at each adaptive step, so Adam has ``converged'' in this process. Below we compare the results of performing this last training with Adam, and with LRSFN using Hessian rank $r = 40$. We used gradient batch size of $2$ for Adam, and gradient batch size of $4$ and Hessian batch size of $2$ for LRSFN. Both optimizers use $\alpha = 10^{-3}$ fixed step sizes. Some results are shown below and they are consistently representative of the pattern seen. 


\begin{figure}[H] 
\begin{subfigure}{0.5\textwidth}
\includegraphics[width = \textwidth]{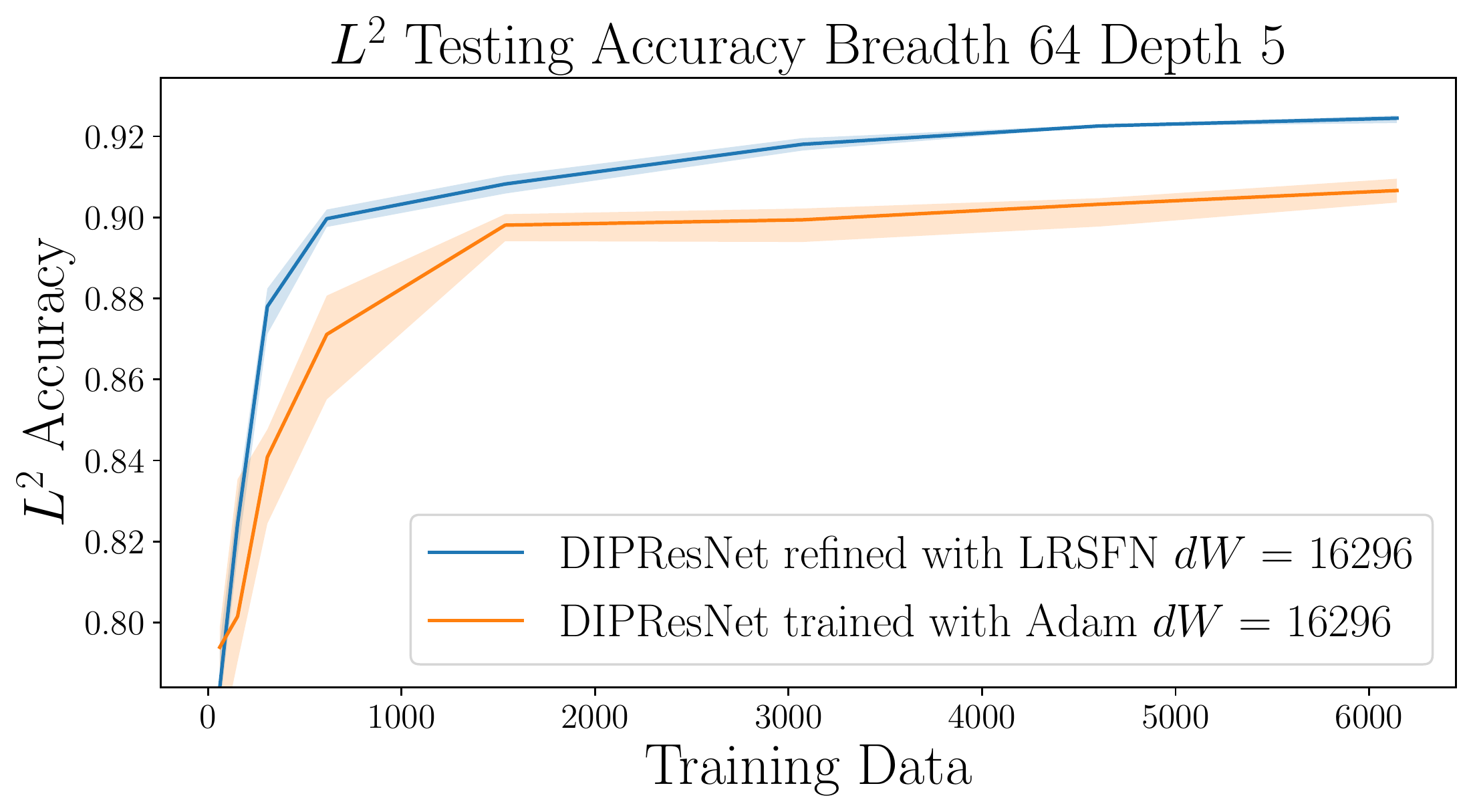}
\end{subfigure}%
\begin{subfigure}{0.5\textwidth}
\includegraphics[width = \textwidth]{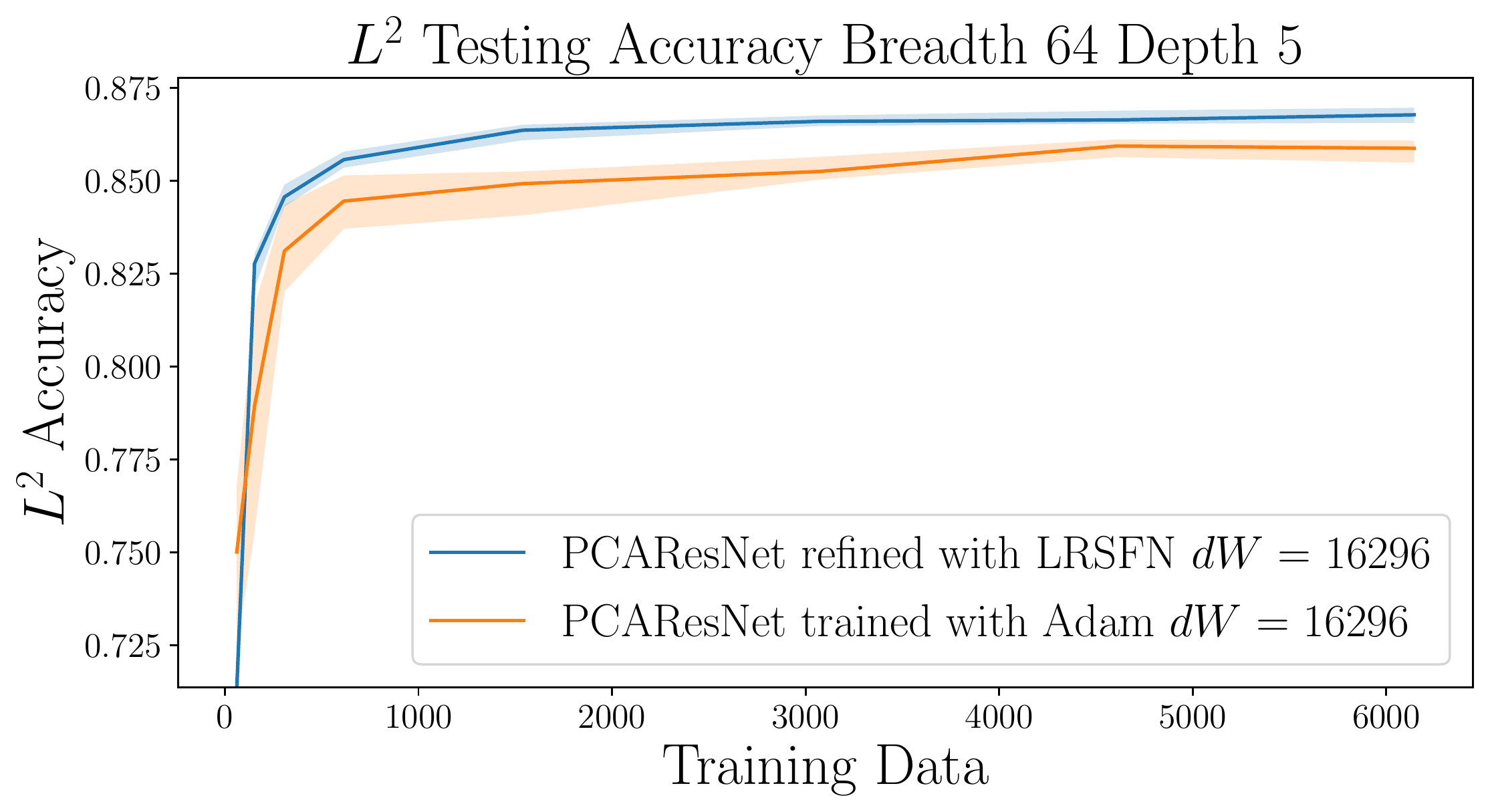}
\end{subfigure}
\caption{Depth 5 Breadth 32 Optimizer Comparison for the Helmholtz Problem.}
\label{depth5_breadth64_opt_comp}
\end{figure}

\begin{figure}[H] 
\begin{subfigure}{0.5\textwidth}
\includegraphics[width = \textwidth]{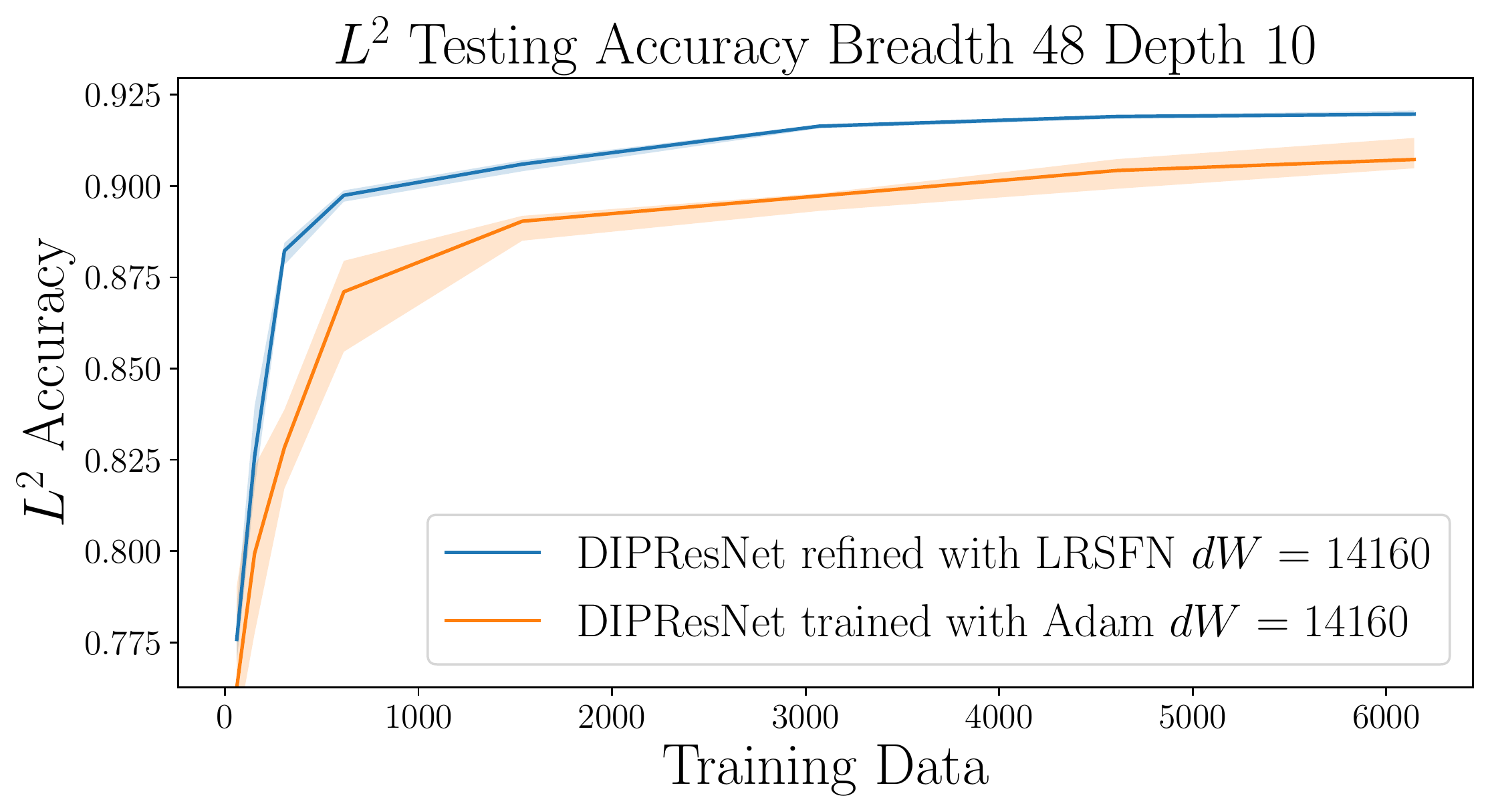}
\end{subfigure}%
\begin{subfigure}{0.5\textwidth}
\includegraphics[width = \textwidth]{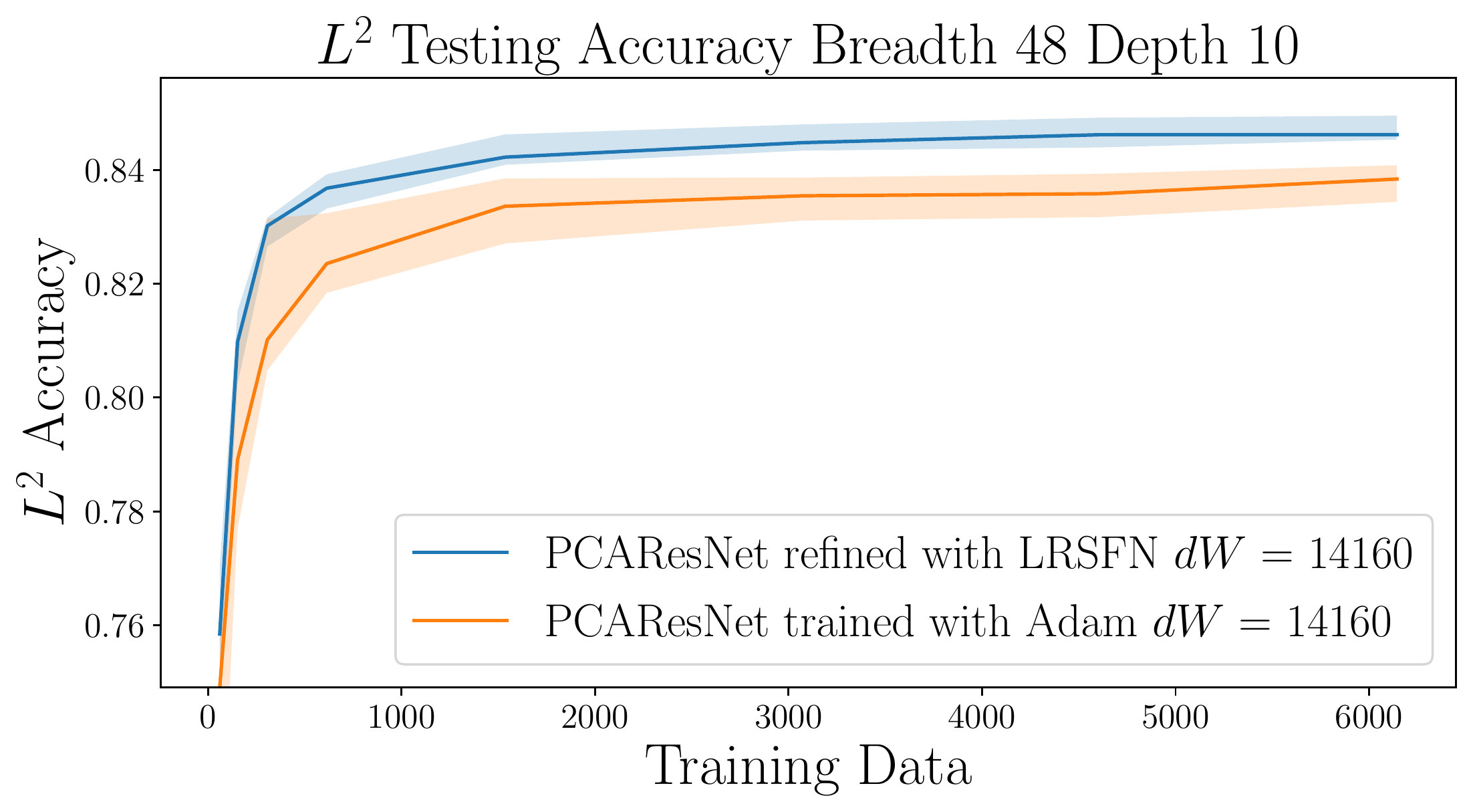}
\end{subfigure}
\caption{Depth 10 Breadth 48 Optimizer Comparison for the Helmholtz Problem.}
\label{depth10_breadth48_opt_comp}
\end{figure}





\bibliographystyle{elsarticle-num}
\bibliography{local}

\end{document}